\documentclass{ws-ijprai}

\usepackage{graphicx,multirow,url,colortbl,hhline,booktabs}
\usepackage{amssymb}
\usepackage{epstopdf}
\usepackage{subfig}
\usepackage[numbers]{natbib}
\usepackage{algpseudocode}
\usepackage{algorithm}
\usepackage{amsmath}
\usepackage{bm}
\usepackage[section]{placeins}
\usepackage{url,overcite}

\newcommand{\Break}{\State \textbf{break} }
\renewcommand{\vec}[1]{\mbox{\boldmath{$#1$}}}
\def\BibTeX{{\rm B\kern-.05em{\sc i\kern-.025em b}\kern-.08em
    T\kern-.1667em\lower.7ex\hbox{E}\kern-.125emX}}

\begin{document}

\algnewcommand{\algorithmicgoto}{\textbf{go to}}%
\algnewcommand{\Goto}[1]{\algorithmicgoto~\ref{#1}}%
\algnewcommand\algorithmicinput{\textbf{Input:}}
\algnewcommand\Input{\item[\algorithmicinput]}
\algnewcommand\algorithmicoutput{\textbf{Output:}}
\algnewcommand\Output{\item[\algorithmicoutput]}

\markboth{R. M. O. Cruz, M. A. Souza, R. Sabourin and G. D. C. Cavalcanti}{On dynamic ensemble selection and data preprocessing for multi-class imbalance learning}

%
\catchline{}{}{}{}{}
%

\title{On dynamic ensemble selection and data preprocessing for multi-class imbalance learning}

%
%

\author{Rafael M. O. Cruz \footnote{Indicates equal contribution}}
\address{Laboratoire d'imagerie, de vision et d'intelligence artificielle\\
\'{E}cole de Technologie Sup\'{e}rieure\\
Montreal, H3C 1K3, Canada\\
\email{rafaelmenelau@gmail.com}}

\author{Mariana A. Souza *}
\address{Centro de Inform\'{a}tica\\
Universidade Federal de Pernambuco\\
Recife, 50.670-420, Brazil\\
\email{mas2@cin.ufpe.br}}

\author{Robert Sabourin}
\address{Laboratoire d'imagerie, de vision et d'intelligence artificielle\\
\'{E}cole de Technologie Sup\'{e}rieure\\
Montreal, H3C 1K3, Canada\\
\email{Robert.Sabourin@etsmtl.ca}}

\author{George D. C. Cavalcanti}
\address{Centro de Inform\'{a}tica\\
Universidade Federal de Pernambuco\\
Recife, 50.670-420, Brazil\\
\email{gdcc@cin.ufpe.br}}

\maketitle


\begin{abstract}
Class-imbalance refers to classification problems in which many more instances are available for certain classes than for others. 
Such imbalanced datasets require special attention because traditional classifiers generally favor the majority class which has a large number of instances. 
Ensemble of classifiers have been reported to yield promising results. However, the majority of ensemble methods applied to imbalanced learning are static ones. 
Moreover, they only deal with binary imbalanced problems. 
Hence, this paper presents an empirical analysis of dynamic selection techniques and data preprocessing methods for dealing with multi-class imbalanced problems. 
We considered five variations of preprocessing  methods and fourteen dynamic selection schemes. 
Our experiments conducted on 26 multi-class imbalanced problems show that the dynamic ensemble improves the AUC and the G-mean as compared to the static ensemble. 
Moreover, data preprocessing plays an important role in such cases.
\end{abstract}

\keywords{Imbalanced learning; multi-class imbalanced; ensemble of classifiers; dynamic classifier selection; data preprocessing.}

\section{Introduction}

Class-imbalance\cite{He09} refers to classification problems in which many more instances are available for certain classes than for others. 
Particularly, in a two-class scenario, one class contains the majority of instances (the \textit{majority class}), while the other (the \textit{minority class}) contains fewer instances. 
Imbalanced datasets may originate from real life problems including the detection of fraudulent bank account transactions\cite{wei2013effective}, telephone calls\cite{akbani2004applying}, biomedical diagnosis\cite{mazurowski2008training}, image retrieval\cite{piras2012synthetic} and so on. 
Due to the under-representation of the minority class, traditional classification algorithms tend to favor the majority class in the learning process. 
This bias leads to a poor performance over the minority class\cite{prati2015class}, which may be an issue since the latter is usually of higher importance than the majority class in many problems, such as in the diagnosis of rare diseases.

One of the biggest challenges in imbalance learning is dealing with multi-class imbalanced problems~\cite{Krawczyk16}. 
Multi-class imbalanced classification is not as well developed as the binary case, with only a few papers handling this issue~\cite{abdi2016combat,fernandez2013analysing,fernandez2011dynamic}. 
It is also considered as a more complicated problem, since the relation among the classes is no longer obvious. 
For instance, one class may be the majority one when compared to some classes, and minority when compared to others. 
Moreover, we may easily lose performance on one class while trying to gain it on another~\cite{fernandez2013analysing}.

A plethora of techniques was designed for addressing imbalanced problems, and they can be classified into one of the following four groups\cite{galar2012review,book-imbalearn}: algorithm-level approaches, data-level approaches, cost-sensitive learning frameworks, and ensemble-based approaches. 
Algorithm-level approaches modify the existing learning algorithms so that they take into account the imbalance between the problem's classes. 
The data-level approaches include sampling-based preprocessing techniques which rebalance the original imbalanced class distribution to reduce its impact in the learning process. 
Cost-sensitive learning frameworks combine both data-level and algorithm-level approaches by assigning misclassification costs to each class and modifying the learning algorithms to incorporate them. 
Lastly, the ensemble-based approaches integrate any of the previous approaches (usually preprocessing techniques) with an ensemble learning algorithm. This work focus on this group of solutions.

As shown in Ref.~\citen{Pastor15}, an ensemble of diverse classifiers can better cope with imbalanced distributions. 
In particular, Dynamic Selection (DS) techniques is seen as an alternative to deal with multi-class imbalanced problems as it explores the local competence of each base classifier according to each new query sample~\cite{CRUZ2018195,Krawczyk16,Roy2018}.  
A few recent works on ensemble-based approaches apply DS for dealing with multi-class imbalanced problems. 
In Ref. \citen{garcia2018dynamic}, the authors proposed a Dynamic Ensemble Selection (DES) technique that combines a preprocessing method based on random balance and a DS scheme which assigns a higher competence to classifiers that correctly label minority class samples in the local region where the query sample is located. 
The proposed preprocessing technique makes use of Random Under-Sampling (RUS), Random Over-Sampling (ROS) and Synthetic Minority Oversampling Technique (SMOTE) for obtaining balanced sets for training the base-classifiers in the pool. 
A DES technique was also proposed in Ref. \citen{krawczyk2018selecting} to deal with multi-class imbalanced problems. 
The clustering-based technique divides the feature space into regions and assigns different weights to the base-classifiers in each area. 
The output of the technique is then given by the weighted response of the local ensemble assigned to the region where the query sample is located. 
The classifiers' weights and the clusters' locations are obtained using an evolutionary scheme with a skew-intensive optimization criterion, with the purpose of reducing the class bias in the responses of the defined local ensembles.

A key factor in dynamic selection is the estimation of the classifiers' competences according to each test sample. 
Usually the estimation of the classifiers competences are based on a set of labelled samples, called the dynamic selection dataset (DSEL). 
However, as reported in Ref.~\citen{cruz2016}, dynamic selection performance is very sensitive to the distribution of samples in DSEL. 
If the distribution of DSEL itself becomes imbalanced, then there is a high probability that the region of competence for a test instance will become lopsided. 
Thus, the dynamic selection algorithms might end up biased towards selecting base classifiers that are experts for the majority class. 
With this in mind, we propose the use of data preprocessing methods for training a pool of classifiers as well as balancing the class distribution in DSEL for the DS techniques.

Hence, in this paper, we perform a study on the application of dynamic selection techniques and data preprocessing for handling with multi-class imbalance. 
Five data preprocessing techniques and nine DS techniques as well as static ensemble combination are considered in our experimental analysis. 
We also evaluate five of the DS techniques within the Frienemy Indecision Region Dynamic Ensemble Selection (FIRE-DES) framework\cite{dayvid}. 
Experiments are conducted using 26 multi-class imbalanced datasets with varying degrees of class imbalance. 
The following research questions are studied in this paper:

\begin{enumerate}
			\item Does data preprocessing play an important role in the performance of dynamic selection techniques?
			\item Which data preprocessing technique is better suited for dynamic and static ensemble combination?
			\item Do dynamic ensembles present better performance than static ensembles?
\end{enumerate}

This paper is organized as follows: Section~\ref{sec:ds} presents the related works on dynamic selection and describes the DCS and DES methods considered in this analysis. 
Data preprocessing techniques for imbalance learning are presented in Section~\ref{sec:preprocessing}. 
Experiments are conducted in Section~\ref{sec:experiments}. 
Conclusion and future works are presented in the last section.

\section{Dynamic selection}
\label{sec:ds}

A dynamic selection (DS) enables the selection of one or more base classifiers from a pool, given a test instance. 
This is based on the assumption that each base classifier is an expert in a different local region in the feature space. 
Therefore, the most competent classifiers should be selected in classifying a given unknown instance. 
The notion of competence is used in DS as a way of selecting, from a pool of classifiers, the best classifiers to classify a given test instance. 

The general process for obtaining a specific Ensemble of Classifiers (EoC) for each query instance can be divided in three steps\cite{CRUZ2018195}: Region of Competence (RoC) definition, competence estimation and selection approach. 
In the first step, the local area in which the query sample is located is obtained. This area is called the Region of Competence (RoC) of the query instance. 
Then, the competence of each classifier in the query sample's RoC is estimated according to a given competence measure in the second step. 
Finally, either the classifier with highest competence level is singled out or an ensemble composed of the most competent classifiers is selected in the last step. 
If the former, the selection approach is a Dynamic Classifier Selection (DCS) scheme. 
If the latter, a Dynamic Ensemble Selection (DES) approach is used. 
Choosing more than one classifier to label the query instance can be advantageous since the risk of selecting an unsuitable one is distributed in DES schemes\cite{Ko08}.
Both these strategies have been studied in recent years, and some papers are available examining them\cite{Britto14,Cruz15}. 

\begin{figure*}[!htbp]
		\centerline{
		\includegraphics[width=1.1\textwidth]{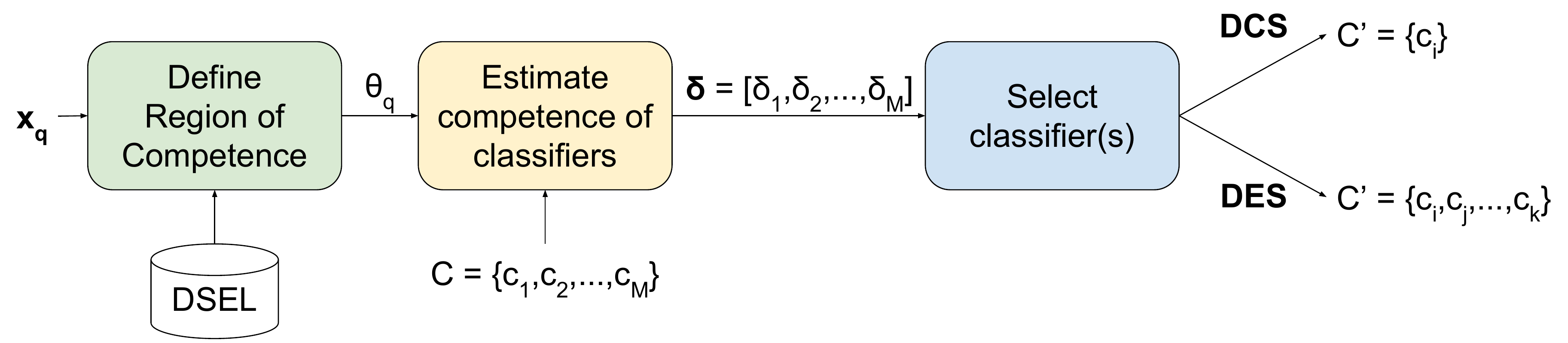}}
		\caption{
		Steps of a DS scheme. 
		DSEL is the dynamic selection dataset, which contains labelled samples, $\mathbf{x_{q}}$ is the query sample, $\theta_{q}$ is the query sample's Region of Competence (RoC), $C$ is the pool of classifiers, $\bm{\delta}$ is the competence vector composed of the estimated competences $\delta_{i}$ of each classifier $c_{i}$ and $C'$ is the resulting EoC of the selection scheme. If the selection approach is DCS, $C'$ will contain only one classifier from $C$. Otherwise, the most competent classifiers in $C$ will be chosen to form the EoC.
		}
		\label{fig:dcs-des}
\end{figure*}

\par Figure \ref{fig:dcs-des} illustrates the usual procedure for dynamically selecting classifiers. 
The query sample $\mathbf{x_{q}}$ and a set of labelled instances called the dynamic selection dataset (DSEL) are used to define the query sample's RoC ($\theta_{q}$). 
The DSEL dataset can be either the training or validation set. 
The RoC $\theta_{q}$ consists of a subset of the DSEL dataset, and it usually contains the closest labelled instances to the query sample obtained using the k-nearest neighbors (KNN) rule. 
Then, the competence of each classifier from the original pool $C$ is estimated over $\theta_{q}$ using a competence measure. 
The estimated competence of classifier $c_{i}$ is denoted as $\delta_{i}$. 
The \textit{competence vector} $\pmb{\delta}$, which contains the competence estimates from all classifiers in $C$, is then used in the selection approach, which can be a DCS or a DES method, to select the EoC $C'$ that will be used to label $\mathbf{x_{q}}$.

To establish the classifiers' competence, given a test instance and the DSEL, the literature reports a number of measures classified into individual-based and group-based measures~\cite{CRUZ2018195}. 
The former only take into account the individual performance of each base classifier, while the latter consider the interaction between the classifiers in the pool for obtaining the competence estimates.   
However, the competence level of the classifiers are calculated differently by different methods within each category. 
That is, the DS techniques may be based on different criteria to estimate the local competence of the base classifiers, such as accuracy\cite{Woods97}, ranking\cite{Sabourin93}, oracle 
information\cite{Ko08}, diversity\cite{desknn} and meta-learning~\cite{Cruz15}, among others. 
Implementation of several DS techniques can be found on DESLib~\cite{deslib}, a dynamic ensemble learning library in Python available at ~\url{https://github.com/Menelau/DESlib}. 
In this paper, we evaluate the three DCS and six DES strategies described next, which are based on various criteria and were among the best performing DS methods according to the experimental analysis conducted in Ref.~\citen{CRUZ2018195}. 
We also evaluate the performance of five of these DS techniques using the FIRE-DES framework\cite{dayvid}, also described in this section, namely the FIRE-LCA, FIRE-MCB, FIRE-KNE, FIRE-KNU and FIRE-DES-KNN.

\subsection*{Modified Classifier Rank (RANK)} 
\par The Modified Classifier Rank\cite{Sabourin93} is a DCS method that exploits ranks of individual classifiers in the pool for each test instance $\mathbf{x_q}$. 
	The competence $\delta_{i}$ of a given classifier $c_{i}$ is estimated as the number of consecutive nearest neighbors $\mathbf{x_{k}} \in \theta_{q}$ it correctly classifies.
	The classifiers are then ranked with respect to their competence level, and the one with highest rank is selected to label the query sample. 
	Algorithm~\ref{alg:rank} describes this procedure.

\begin{algorithm}[!h]
\centering
\small
\begin{algorithmic}[1]
\Input $DSEL, C$ \Comment{DSEL dataset and pool of classifiers}
\Input $\mathbf{x_q}, K$ \Comment{Query sample and RoC size}
\Output $c_k$ \Comment{The highest ranked classifier}
\State $\theta_q \gets KNN(DSEL,K)$ \Comment{RoC definition}
\For {every $c_{i}$ in $C$} 
	\State $ \delta_i \gets 0$ 
	\For {every $\mathbf{x_{j}}$ in $\theta_q$} 
		\If {$c_i$ correctly labels $\mathbf{x_{j}}$}
			\State $ \delta_i \gets \delta_i + 1$ 
		\Else
			\Break
		\EndIf
	\EndFor 
\EndFor 
\State {Rank the classifiers in $C$ according to $\pmb{\delta}$}\\
\Return {the highest ranked classifier $c_k$}
\end{algorithmic}
\caption{RANK method.}
\label{alg:rank}
\end{algorithm}	
	
\subsection*{Local Class Accuracy (LCA)} 
\par The Local Class Accuracy method\cite{Woods97} estimates the classifiers' competence as their accuracy over the query sample's local region, taking into account the label they assigned to the test instance $\mathbf{x_q}$, as shown in Algorithm~\ref{alg:lca}.
	Thus, for a given classifier $c_i$, which assigned the label $\omega_l$ to the query sample, its competence $\delta_{i}$ is defined as the percentage of correctly classified instances among the ones in the region of competence $\theta_{q}$ that belong to class $\omega_{l}$.
	The classifier presenting the highest competence is used for the classification of the query sample.
	
\begin{algorithm}[!h]
\centering
\small
\begin{algorithmic}[1]
\Input $DSEL, C$ \Comment{DSEL dataset and pool of classifiers}
\Input $\mathbf{x_q}, K$ \Comment{Query sample and RoC size}
\Output $c_k$ \Comment{The most competent classifier}
\State $\theta_q \gets KNN(DSEL,K)$ \Comment{RoC definition}
\For {every $c_{i}$ in $C$} 
	\State $ \delta_i \gets 0$ 
	\State $\omega_l \gets c_i(\mathbf{x_q})$ \Comment{Predict output of the query sample}
	\State $ \theta_q' \gets \{\mathbf{x_{j}} \in \theta_q | \mathbf{x_{j}}$ belongs to class $\omega_l\}$ \Comment{Select neighbors from class $\omega_l$}
	\For {every $\mathbf{x_{j}}$ in $\theta_q'$} \Comment{Compute accuracy within $\theta_q'$}
		\If {$c_i$ correctly labels $\mathbf{x_{j}}$}
			\State $ \delta_i \gets \delta_i + 1$ 
		\EndIf
	\EndFor 
	\State $\delta_i \gets {\delta_i}/{|\theta_q'|}$ \Comment{Calculate proportion of correctly classified samples in $\theta_q'$}
\EndFor 
\State {Select the classifier $c_k$ in $C$ with highest $\delta_k$}\\
\Return {the most competent classifier $c_k$}
\end{algorithmic}
\caption{LCA method.}
\label{alg:lca}
\end{algorithm}

\subsection*{Multiple Classifier Behavior (MCB)} 
The Multiple Classifier Behavior\cite{Giacinto01} is a DCS technique based on both accuracy and behavior of the classifiers. 
	The latter is a characteristic that comes from the \textit{decision space}, which relates to the responses given by the base-classifiers in the pool. 
	A sample $\mathbf{x_i}$ in the feature space may be represented as an instance in the decision space by its output profile $\mathbf{u_i}$, which consists of all base-classifiers' predictions for that sample. 
	The decision space may be used to various ends, from RoC definition, as in the case of MCB, to even learning the behavior of the classifiers in the pool\cite{icprai-metalearn,Cruz15}.

\par Algorithm~\ref{alg:mcb} describes the selection procedure by the MCB method. 
	Given an unknown instance $\mathbf{x_q}$, the output profile ($\mathbf{u_k}$) of each of its neighbors $\mathbf{x_k} \in \theta_{q}$ is first obtained.
	Then, the output profile of the test instance ($\mathbf{u_q}$) is also obtained and compared to the output profile of each of its neighbors using a similarity measure (Eq. (\ref{eq:mcb})) defined as the proportion of equal corresponding coordinate values between the two profiles. 

\begin{equation} \label{eq:mcb}
\centering
\begin{split}
	Sim(\mathbf{u_i},\mathbf{u_j}) = \frac{1}{M}  \sum_{k = 1}^{M} T_k(\mathbf{u_{i}},\mathbf{u_{j}}),
	\\
	T_k(\mathbf{u_{i}},\mathbf{u_{j}}) = \begin{cases}
						0, \textnormal{if } u_{i,k} \neq u_{j,k} \\
						1, \textnormal{if } u_{i,k} = u_{j,k}
						\end{cases}
\end{split}
\end{equation}	

\begin{algorithm}[!h]
\centering
\small
\begin{algorithmic}[1]
\Input $DSEL, C$ \Comment{DSEL dataset and pool of classifiers}
\Input $\mathbf{x_q}, K$ \Comment{Query sample and RoC size}
\Input $t_s, t_c$ \Comment{Similarity and competence thresholds}
\Output $c_k$ \Comment{The most competent classifier}
\State $ \mathbf{u_q} \gets C(\mathbf{x_q}) $ \Comment{Compute the responses of all classifiers (i.e., the output profile) of $\mathbf{x_{q}}$}
\State $\theta_q \gets KNN(DSEL,K)$ \Comment{RoC definition}
\State $\theta_q' \gets \{\}$ \Comment{New RoC}
\For {every $\mathbf{x_{j}}$ in $\theta_q$}
	\State $ \mathbf{u_j} \gets C(\mathbf{x_j})$ \Comment{Compute the output profile of $\mathbf{x_{j}}$}
	\If {$Sim(\mathbf{u_q},\mathbf{u_j}) > t_s$}		\Comment{Compare the output profiles (Eq. (\ref{eq:mcb}))}
		\State $\theta_q' \gets \theta_q' \cup \mathbf{x_{j}}$ 
	\EndIf
\EndFor 
\State $ \delta_i \gets 0$ 
\For {every $\mathbf{x_{j}}$ in $\theta_q'$} \Comment{Compute accuracy within $\theta_q'$}
		\If {$c_i$ correctly labels $\mathbf{x_{j}}$}
			\State $ \delta_i \gets \delta_i + 1$ 
		\EndIf
	\State $\delta_i \gets {\delta_i}/{|\theta_q|}$ \Comment{Calculate proportion of correctly classified samples in $\theta_q$}
\EndFor
\State {Select the classifier $c_k$ in $C$ with highest $\delta_k$}
\If {$\delta_k$ is greater by $t_c$ than all other competences in $\pmb{\delta}$}
	\State {\textbf{return} the most competent classifier $c_k$}
\Else
	\State {\textbf{return} $C$}
\EndIf
\end{algorithmic}
\caption{MCB method.}
\label{alg:mcb}
\end{algorithm}	
	
	The instances with output profiles similar to the query sample's, according to a pre-defined threshold, remain in the region of competence, while the rest is removed from $\theta_{q}$. 
	Then, the competence $\delta_{i}$ of classifier $c_{i}$ is estimated as its local accuracy over the new region of competence, and if the difference between the most accurate classifier and the second most accurate surpasses a second threshold, the former is selected to label the query sample. 
	Otherwise, the majority voting of all classifiers is used for classification.

\subsection*{KNORA-Eliminate (KNE)}
\par The KNORA-Eliminate technique ~\cite{Ko08} explores the concept of Oracle, which is the upper limit of a DCS technique. 
	Given the region of competence $\theta_q$, only the classifiers that correctly recognize all samples belonging to the region of competence are selected. 
	In other words, all classifiers that achieved a 100\% accuracy in this region (i.e., that are local Oracles) are selected to compose the ensemble of classifiers. 
	Then, the decisions of the selected base classifiers are aggregated using the majority voting rule. 
	If no base classifier is selected, the size of the region of competence is reduced by one, and the search for the competent classifiers is restarted. 
	This procedure is described in Algorithm~\ref{alg:kne}.
	
\begin{algorithm}[!h]
\centering
\small
\begin{algorithmic}[1]
\Input $DSEL, C$ \Comment{DSEL dataset and pool of classifiers}
\Input $\mathbf{x_q}, K$ \Comment{Query sample and RoC size}
\Output $C'$ \Comment{EoC}
\State $\theta_q \gets KNN(DSEL,K)$ \Comment{RoC definition}
\State $\theta_q' \gets \theta_q$ \Comment{New RoC is initially the original RoC}
\While {$\theta_q'$ is not empty}
	\For {every $c_{i}$ in $C$} 
		\State $ \delta_i \gets 0$ 
		\For {every $\mathbf{x_{j}}$ in $\theta_q'$} \Comment{Compute accuracy within $\theta_q'$}
			\If {$c_i$ correctly labels $\mathbf{x_{j}}$}
				\State $ \delta_i \gets \delta_i + 1$ 
			\EndIf
		\EndFor 	
	\EndFor 
	\If {there are $c_k$ with $\delta_k = |\theta_q'|$}
		\State $C' \gets C' \cup c_{k}$ \Comment{Add perfectly accurate classifiers to $C'$}
	\Else
		\State {remove the most distant sample $\mathbf{x_j}$ from $\theta_q'$} \Comment{Reduce the new RoC $\theta_q'$ by one}
	\EndIf
\EndWhile
\If {$\theta_q'$ is not empty}
	\State {\textbf{return} $C'$}
\Else
	\State {\textbf{return} $C$}
\EndIf
\end{algorithmic}
\caption{KNE method.}
\label{alg:kne}
\end{algorithm}	
	
\subsection*{KNORA-Union (KNU)}

\par The KNORA-Union technique~\cite{Ko08} selects all classifiers that are able to correctly recognize at least one sample in the region of competence $\theta_q$. 
	The competence $\delta_{i}$ of a given classifier $c_{i}$ is estimated as the number of samples in $\theta_q$ for which it predicted the correct label. 
	The method then selects all classifiers with competence level above zero. 
	The KNU selection scheme is shown in Algorithm~\ref{alg:knu}.
	The responses of the selected classifiers are combined using a majority voting scheme which considers that a base classifier can vote more than once when it correctly classifies more than one instance in the region of competence.
	For instance, if a given base classifier $c_{i}$ predicts the correct label for three samples belonging to the region of competence, it gains three votes for the majority voting scheme. The votes collected by all base classifiers are aggregated to obtain the ensemble decision. So, in addition to the selected EoC, the KNU technique returns the competence vector $\pmb{\delta}$ to be used by the aggregation scheme, as shown in Algorithm~\ref{alg:knu}.
	
\begin{algorithm}[!h]
\centering
\small
\begin{algorithmic}[1]
\Input $DSEL, C$ \Comment{DSEL dataset and pool of classifiers}
\Input $\mathbf{x_q}, K$ \Comment{Query sample and RoC size}
\Output $C',\pmb{\delta}$ \Comment{EoC and competence estimates}
\State $\theta_q \gets KNN(DSEL,K)$ \Comment{RoC definition}
\For {every $c_{i}$ in $C$} 
	\State $ \delta_i \gets 0$ 
	\For {every $\mathbf{x_{j}}$ in $\theta_q$} \Comment{Compute accuracy within $\theta_q$}
		\If {$c_i$ correctly labels $\mathbf{x_{j}}$}
			\State $ \delta_i \gets \delta_i + 1$ 
		\EndIf
	\EndFor 
	\If {$\delta_i > 0$}
		\State $C' \gets C' \cup c_i$ \Comment{Add classifiers with competence above zero}
	\EndIf	
\EndFor \\
\Return {the EoC $C'$ and the competence estimates $\pmb{\delta}$}
\end{algorithmic}
\caption{KNU method.}
\label{alg:knu}
\end{algorithm}	
	
\subsection*{DES-KNN} 
The DES-KNN\cite{desknn} technique relies on both an individual-based measure and a group-based measure, namely local accuracy and diversity, to estimate the classifiers' competence. 
	Algorithm~\ref{alg:desknn} describes its selection scheme. Firstly, the individual local accuracy of each classifier $c_i$ is estimated over the query instance's region of competence $\theta_q$, obtained using the KNN. 
	Then, the $N$ classifiers with highest local accuracies are pre-selected, with $N$ being a pre-defined parameter. 
	The pre-selected classifiers are then ranked again based on their diversity using a diversity measure so that the $J$ most diverse ones among them, with $J$ also being pre-set, are used for classifying the query sample $\mathbf{x_q}$. 
	The diversity measure used was the Double-Fault measure\cite{giacinto2001design}, since it provided the highest correlation with accuracy in Ref.~\citen{shipp2002relationships}.
	
\begin{algorithm}[!h]
\centering
\small
\begin{algorithmic}[1]
\Input $DSEL, C$ \Comment{DSEL dataset and pool of classifiers}
\Input $\mathbf{x_q}, K$ \Comment{Query sample and RoC size}
\Input $N, J$ \Comment{Parameters}
\Output $C'$ \Comment{EoC}
\State $\theta_q \gets KNN(DSEL,K)$ \Comment{RoC definition}
\For {every $c_{i}$ in $C$} 
	\State $ \delta_i \gets 0$ 
	\For {every $\mathbf{x_{j}}$ in $\theta_q$} \Comment{Compute accuracy within $\theta_q$}
		\If {$c_i$ correctly labels $\mathbf{x_{j}}$}
			\State $ \delta_i \gets \delta_i + 1$ 
		\EndIf
	\EndFor 
\EndFor
\For {every $c_{i}$ in $C$} 
	\For {every $c_{j}$ in $C$} 
		\If $i \neq j$ 
			\State $D_{i,j} \gets DoubleFault(c_i,c_j)$ \Comment{Compute the diversity between $c_i$ e $c_j$}
		\EndIf
	\EndFor
\EndFor
\State {Rank the classifiers in $C$ according to $\pmb{\delta}$}
\State $C'' \gets$ {the $N$ most accurate classifiers in $C$}
\State {Rank the classifiers in $C''$ according to $D$}
\State $C' \gets$ {the $j$ most diverse classifiers in $C''$}\\
\Return {the EoC $C'$}
\end{algorithmic}
\caption{DES-KNN method.}
\label{alg:desknn}
\end{algorithm}	
	
\subsection*{Dynamic Ensemble Selection Performance (DES-P)} 
The Dynamic Ensemble Selection Performance technique\cite{WoloszynskiKPS12} estimates the competence of each classifier as the difference between its local accuracy over the region of competence $\theta_q$ and the performance of the random classifier, that is, the classification model that randomly chooses a class with equal probabilities. 
	The performance of the random classifier is defined by $RC = 1/L$, with $L$ being the number of classes in the problem. 
	Thus, for a given classifier $c_i$, its competence $\delta_i$ is estimated by Eq. (\ref{eq:desp}), with $\hat{P}(c_i|\theta_q)$ being the classifier's local accuracy over the region of competence.
	\begin{equation}\label{eq:desp}
		\delta_i = \hat{P}(c_i|\theta_q) - \dfrac{1}{L}
	\end{equation}		
	After estimating the classifiers' competences, the technique selects the ones with competence level above zero, that is, the classifiers with local accuracy greater than the random classifier's. 
	If no classifier meets this requirement, the entire pool is used for classifying the query sample. 
	The selection procedure of the DES-P method is shown in Algorithm~\ref{alg:desp}.
	
\begin{algorithm}[!h]
\centering
\small
\begin{algorithmic}[1]
\Input $DSEL, L, C$ \Comment{DSEL dataset, number of classes and pool of classifiers}
\Input $\mathbf{x_q}, K$ \Comment{Query sample and RoC size}
\Output $C'$ \Comment{EoC}
\State $\theta_q \gets KNN(DSEL,K)$ \Comment{RoC definition}
\State $RC \gets 1/L$ \Comment{Compute the performance of the random classifier}
\For {every $c_{i}$ in $C$} 
	\State $ \delta_i \gets 0$ 
	\For {every $\mathbf{x_{j}}$ in $\theta_q$} \Comment{Compute accuracy within $\theta_q$}
		\If {$c_i$ correctly labels $\mathbf{x_{j}}$}
			\State $ \delta_i \gets \delta_i + 1/|\theta_q|$ 
		\EndIf
	\EndFor 	
	\State $ \delta_i \gets \delta_i - RC$ \Comment{Compute competence level (as in Eq. (\ref{eq:desp}))}
\EndFor
\If {there are $c_k$ with $\delta_k > 0$}
	\State $C' \gets C' \cup c_{k}$ \Comment{Add classifiers with competence above RC to $C'$}
	\State {\textbf{return} $C'$}
\Else
	\State {\textbf{return} $C$}
\EndIf
\end{algorithmic}
\caption{DES-P method.}
\label{alg:desp}
\end{algorithm}	
	
\subsection*{Randomized Reference Classifier (DES-RRC)} 

The Randomized Reference Classifier\cite{woloszynski2011probabilistic} is a probability-based DS technique which makes use of the random reference classifier (RRC) from Ref. \citen{Woloszynski2010AMO} for obtaining the classifiers' competence. 
	More specifically, the competence level $\delta_i$ of a given classifier $c_i$ is defined as the weighted sum of its source of competence $C_{src}$, calculated using the probability of correct classification of a RRC, over all validation instances (Eq. (\ref{eq:rrc})). 
	The weights are given by a Gaussian potential function, shown in Eq. (\ref{eq:gauss}), whose value decreases with the increase of the Euclidean distance between the query instance and the validation sample. 
	\begin{equation}\label{eq:rrc}
		\delta_i = \sum_{\mathbf{x_k} \in DSEL} C_{src}K(\mathbf{x_q},\mathbf{x_k})
	\end{equation}	
	\begin{equation}\label{eq:gauss}
		K(\mathbf{x_q},\mathbf{x_k}) = exp(-d(\mathbf{x_q},\mathbf{x_k})^2)
	\end{equation}
	The classifiers with competence estimate greater than the probability of random classification are selected and used for classifying the unknown sample.
	

\subsection*{META-DES} 
\par The META-DES\cite{Cruz15} is a framework based on the premise that the dynamic ensemble selection problem can be considered as a meta-problem\cite{cruz2014meta}, which uses different criteria regarding the behavior of $c_i$ to decide if it is competent or not to label the query sample $\mathbf{x_q}$. The meta-problem is defined as follows:
	\begin{itemize}
	\item The \textbf{meta-classes} are either ``competent" (1) or ``incompetent" (0) to classify $\mathbf{x_q}$.
	\item Each set of \textbf{meta-features} $f_i$ corresponds to a different criterion for measuring the level of competence of a base classifier.
	\item The meta-features are encoded into a meta-features vector $v_{i,q}$.
	\item A \textbf{meta-classifier} $\lambda$ is trained based on the meta-features $v_{i,q}$ to predict whether or not $c_i$ will achieve the correct prediction for $\mathbf{x_q}$, i.e., if it is competent enough to classify $\mathbf{x_q}$.
	\end{itemize}
	Thus, a meta-classifier $\lambda$ is trained and used to decide whether or not each classifier in the pool is competent enough to classify a given query sample $\mathbf{x_q}$. 
	During the meta-training phase of the framework, the meta-features are extracted from each sample in the training and DSEL sets. 
	The set of meta-features may include different criteria for measuring a classifier's competence, such as local accuracy, output profile, posterior probability, and so on. 
	The meta-classifier $\lambda$ is then trained using the extracted meta-features, in order to learn the classifier selection rule. 
	During the generalization phase, the meta-features of a given unknown sample are extracted and used as input to the meta-classifier, which estimates the competence level $\delta_i$ of each classifier $c_i$ in the pool, deeming them as competent or not for classifying that specific sample.
	The competent classifiers are then selected to perform the classification task using majority voting.

\subsection*{FIRE-DES} 
\par
\begin{figure}[h]

			\centering
			\includegraphics[width=0.4\textwidth]{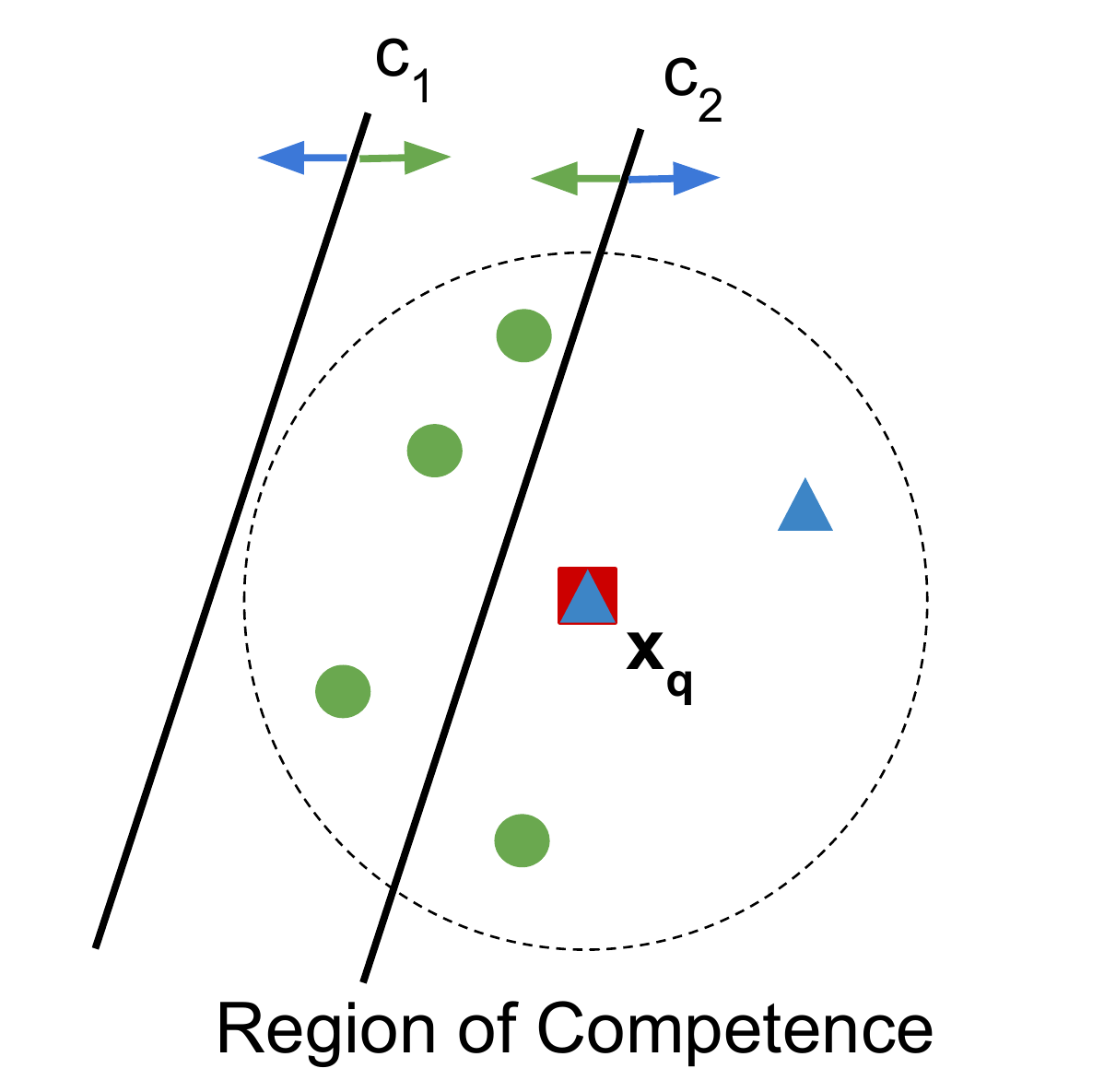}
			\caption{Example in which a query sample $\mathbf{x_q}$, which belongs to the blue class, is located in an indecision region. While classifier $c_1$ only recognizes the instances from the majority class (green), $c_2$ crosses the RoC and correctly labels instances from both classes.}
			\label{fig:ex-fire}
		\end{figure}

 Though not a DS technique, the FIRE-DES framework exploits the local competence of the classifiers by performing a dynamic ensemble pruning on the original pool of classifiers $C$ prior to the use of a DS scheme. 
The idea behind the FIRE-DES is that a number of DS techniques may still select incompetent classifiers for labelling unknown instances located in \textit{indecision regions}, that is, areas near the class borders. 
This issue is illustrated in Figure \ref{fig:ex-fire}, in which $\mathbf{x_q}$ is a query sample located in an indecision region and $c_1$ and $c_2$ are two classifiers from the pool. 

\par In this example, both classifiers correctly label 4 out of 5 neighbors of $\mathbf{x_q}$. 
However, while $c_1$ only recognizes the green class in the query sample's RoC, $c_2$ is able to predict different labels in this region because it ``crosses" the RoC. 
Moreover, $c_2$ recognizes the local border in the RoC since it correctly labels at least one pair of \textit{frienemies}, that is, samples from different classes in it. 
Thus, it would be preferable to select $c_2$ instead of $c_1$ to label $\mathbf{x_q}$. 
Still, as both classifiers have the same local accuracy, a DS technique could select $c_1$ instead of $c_2$, and thus misclassify $\mathbf{x_q}$. 
This issue may be intensified for highly imbalanced problems, since the DS techniques may select classifiers that simply recognize well the majority class in the RoC, which would yield them a high local accuracy estimate regardless of their recognition of a local border near the query sample. 

\begin{algorithm}[!h]
\centering
\small
\begin{algorithmic}[1]
\Input $DSEL, C$ \Comment{DSEL dataset and pool of classifiers}
\Input $\mathbf{x_q}, K$ \Comment{Query sample and RoC size}
\Output $C''$ \Comment{Subset of classifiers}
\State $\theta_q \gets KNN(DSEL,K)$ \Comment{RoC definition}
\If {there is more than one class in $\theta_q$}
	\State $F \gets \{\}$ \Comment{Set of frienemies in $\theta_q$}
	\For {every $\mathbf{x_{i}}$ in $\theta_q$} 
		\For {every $\mathbf{x_{j}}$ in $\theta_q$} 
			\If {$\mathbf{x_{i}}$ and $\mathbf{x_{j}}$ belong to different classes}
				\State $F \gets F \cup (\mathbf{x_{i}},\mathbf{x_{j}})$ \Comment{Single out the pair of frienemies}
			\EndIf
		\EndFor
	\EndFor
	\State $C'' \gets \{\}$
	\For {every $c_i$ in $C$}
		\If {$c_i$ correctly labels one pair of samples in $F$}
			\State $C'' \gets C'' \cup c_i$ \Comment{Add the classifiers that label a pair of frienemies correctly}
		\EndIf
	\EndFor
	\State {\textbf{return} $C''$}
\Else
	\State {\textbf{return} $C$}
\EndIf
\end{algorithmic}
\caption{DFP method.}
\label{alg:dfp}
\end{algorithm}	

\par Thus, in order to avoid the selection of classifiers that do not ``cross", that is, that do not recognize more than one class in the indecision region where $\mathbf{x_q}$ is, the FIRE-DES dynamically removes such classifiers from the original pool before proceeding with the DS technique execution. 
To that end, the FIRE-DES framework uses the Dynamic Frienemy Pruning (DFP) method (Algorithm~\ref{alg:dfp}), which performs an online pruning of the pool of classifiers based on the neighborhood of each test sample. 
Thus, if an unknown instance's RoC $\theta_q$ contains more than one class, the DFP pre-selects the classifiers from the pool that correctly label at least one pair of frienemies. 
The DS technique is then executed using the pre-selected pool yielded by the DFP instead of the original unpruned pool. 
The FIRE-DES scheme was designed for two-class problems and it yielded a significant improvement in the performance of most DS techniques, specially for highly imbalanced binary problems\cite{dayvid}.

\vspace*{0.3in}

\par Nevertheless, a crucial aspect in the performance of the dynamic selection techniques is the distribution of the dynamic selection dataset (DSEL), as the local competence of the base classifiers are estimated based on this set. 
Hence, preprocessing techniques can really benefit DS techniques as they can be employed to edit the distribution of DSEL, prior to performing dynamic selection.

\section{Data preprocessing}
\label{sec:preprocessing}
		
		Changing the distribution of the training data to compensate for poor representativeness of the minority class is an effective solution for imbalanced problems, and a plethora of methods are available in this regards. Branco et al.~\cite{Branco16} divided such methods into three categories, namely, stratified sampling, synthesizing new data, and combinations of the two previous methods. While the complete taxonomy is available in Ref.~\citen{Branco16}, we will center our attention on the methods that have been used together with ensemble learning~\cite{Pastor15}. 
		
		One important category is under-sampling, which removes instances from the majority class to balance the distribution.
		Random under-sampling (RUS)\cite{Barandela03} is one such method. RUS has been coupled with boosting (RUSBoost)\cite{Seiffert10} and with Bagging\cite{Barandela03}. These combined techniques have been applied to several inherently imbalanced classification problems, such as sound event detection \cite{yang2018sound}, phishing detection \cite{gutierrez2018learning}, software defect prediction \cite{song2018comprehensive}, detection of cerebral microbleds \cite{ateeq2018ensemble} and breast cancer cytological malignancy grading \cite{icprai-rusboost}, among others. A major drawback of RUS is that it can discard potentially useful data which can be a problem when using dynamic selection approaches.
		
		The other strategy is the generation of new synthetic data. Synthesizing new instances has several known advantages\cite{Chawla02}, and a wide number of proposals are available for building new synthetic examples. In this context, a famous method that uses interpolation to generate new instances is SMOTE~\cite{Chawla02}. SMOTE over-samples the minority class by generating new synthetic data. A number of methods have been developed based on the principle of SMOTE, such as, Borderline-SMOTE\cite{Han05}, ADASYN\cite{He08}, RAMO\cite{Chen10} and Random balance\cite{Pastor15RB}. Furthermore, Garcia et al.~\cite{Garcia12} observed that over-sampling consistently outperforms under-sampling for strongly imbalanced datasets. 
		
		Hence, in this work we considered three over-sampling techniques. Similar to Ref.~\citen{abdi2016combat}, the class with the highest number of examples is considered the majority class, while all others are considered minority classes. Then, the over-sampling techniques are applied to generate synthetic samples for each minority class.

\subsection*{Synthetic Minority Over-sampling Technique (SMOTE)} 
The Synthetic Minority Over-sampling Technique\cite{Chawla02} creates artificial instances for the minority class by interpolating several samples that are within a defined neighborhood. The general idea of the process is as follows: Let $\mathbf{x_i}$ be a randomly selected instance from the minority class. To create an artificial instance from $\mathbf{x_i}$, SMOTE first isolates the k-nearest neighbors of $\mathbf{x_i}$, from the minority class. Afterwards, it randomly selects one neighbor and randomly generates a synthetic example along the imaginary line connecting $\mathbf{x_i}$ and the selected neighbor. 
The complete SMOTE procedure is shown in Algorithm~\ref{alg:smote}, in which T, N and K are the three input parameters denoting the number of minority class samples, the amount of oversampling in terms of percentage and the number of nearest neighbors, respectively.

\begin{algorithm}[!h]
\centering
\small
\begin{algorithmic}[1]
\Input $S_P$ \Comment{Set of minority class samples} 
\Input $T, N, K$ \Comment{No. of minority class samples, amount of oversampling and neighborhood size}
\Output $S_S$ \Comment{Set of synthetic samples (of size $(N/100)*T$)}
\If {$N < 100$}
	\State Randomize the $T$ minority class samples
	\State $T \gets (N/100)*T$
	\State $N \gets 100$
\EndIf
\State $N \gets \lfloor N/100\rfloor$ \Comment{$N$ is assumed to be in integral multiples
of 100}
\State $S_s \gets \{\}$
\For {$i$ in $[1,T]$}
	\State $\theta \gets KNN(S_P,K)$ \Comment{Get nearest neighbors of $\mathbf{x_i}$}
	\While {$N > 0$}
		\State {Randomly select a $\mathbf{x_k} \in \theta$}
		\For {each feature $j$} 
			\State $d \gets x_{k,j} - x_{i,j}$ \Comment{Difference in attribute $j$}
			\State $g \gets random([0,1])$ \Comment{Randomized gap}
			\State $x_{i,j}' \gets x_{i,j} + g*d$ \Comment{New value in attribute $j$}
		\EndFor
		\State $S_S \gets S_S \cup \mathbf{x_i'}$
		\State $N \gets N - 1$	
	\EndWhile
\EndFor
\State {\textbf{return} $S_S$}
\end{algorithmic}
\caption{SMOTE procedure.}
\label{alg:smote}
\end{algorithm}	
			
\subsection*{Ranked Minority Over-sampling (RAMO)} 
The Ranked Minority Over-sampling\cite{Chen10} method performs a sampling of the minority class according to a probability distribution, followed by the creation of synthetic instances. The RAMO process works as follows: For each instance $\mathbf{{x}_i}$ in the minority class, its $k_1$ nearest neighbors ($k_1$ is a user defined neighborhood size) from the whole dataset are isolated. The weight $r_i$ of $\mathbf{{x}_i}$ is defined as:
			\begin{equation}\label{eqn:RAMO_Weight}
			r_i = \frac{1}{1+exp(-\alpha.\delta_i)},
			\end{equation}
			where $\delta_i$ is the number of majority cases in the k-nearest neighborhood. Evidently, an instance with a large weight indicates that it is surrounded by majority class samples, and thus difficult to classify.
			
			After determining all weights, the minority class is sampled using these weights to get a sampling minority dataset $G$. The synthetic samples are generated for each instance in $G$ by using SMOTE on $k_2$ nearest neighbors where $k_2$ is a user-defined neighborhood size. 
			
\subsection*{Random Balance (RB)}
The Random Balance method\cite{Pastor15RB} relies on the amount of under-sampling and over-sampling that is problem specific and that has a significant influence on the performance of the classifier concerned. RB maintains the size of the dataset, but varies the proportion of the majority and minority classes, using a random ratio. This includes the case where the minority class is over represented and the imbalance ratio is inverted. Thus, repeated applications of RB produce datasets having a large imbalance ratio variability, which promotes diversity~\cite{Pastor15RB}. SMOTE and random under-sampling are used to respectively increase or reduce the size of the classes to achieve the desired ratios. 

\begin{algorithm}[!h]
\centering
\small
\begin{algorithmic}[1]
\Input $S,S_P,S_N$ \Comment{Original dataset, set of minority class samples and set of majority class samples} 
\Input $K$ \Comment{Neighborhood size of SMOTE}
\Output $S'$ \Comment{Set of generated samples}
\State $totalSize \gets |S|$ 
\State $majSize \gets |S_P|$
\State $minSize \gets |S_N|$
\State $newMajSize \gets random(2,totalSize - 2)$ \Comment{Randomly obtain the new class sizes}
\State $newMinSize \gets totalSize - newMajSize$
\State $S' \gets \{\}$
\If {$newMajSize < majSize$}
	\State $S' \gets S_P$ \Comment{Include all minority samples}
	\State $S' \gets S' \cup RUS(S_N, newMajSize)$ \Comment{Apply RUS on the majority class}
	\State $S' \gets S' \cup SMOTE(S_P,(newMinSize-minSize)/|S_P|,K)$ \Comment{Generate minority class samples using SMOTE}
\Else
	\State $S' \gets S_N$ \Comment{Include all majority samples}
	\State $S' \gets S' \cup RUS(S_P,newMinSize)$ \Comment{Apply RUS on the minority class}
	\State $S' \gets S' \cup SMOTE(S_N,(newMajSize-majSize)/|S_N|,K)$ \Comment{Generate majority class samples using SMOTE}
\EndIf
\State {\textbf{return} $S'$}
\end{algorithmic}
\caption{RB procedure.}
\label{alg:rb}
\end{algorithm}	

The RB procedure is described in Algorithm~\ref{alg:rb}, in which $S$ is the dataset, $S_P$ is the minority class set and $S_N$ is the majority class set. Firstly, the new size of the majority class ($newMajSize$) is obtained by generating a random number between $2$ and $|S|-2$, which leaves the minority class with size $|S| - newMajSize$. Then, if the new majority class size is smaller than its original size $|S_N|$, the majority class $S'_N$ is created by RUS the original $S_N$ so that the final size $|S'_N| = newMajSize$. Consequently, the new minority class $S'_P$ is obtained from $S_P$ using SMOTE to create $newMinSize-|S_P|$ artificial instances. Otherwise, the new minority class $S'_P$ is obtained by RUS the original minority class $S_P$ until $|S'_P| = newMinSize$, while the remaining $newMajSize-|S_N|$ samples from the majority class are created using SMOTE, so that the final size of $|S'_N|$ is $newMajSize$.

\section{Experiments}
\label{sec:experiments}
\subsection{Datasets}

				A total of 26 multi-class imbalanced datasets taken from the Keel repository~\cite{Fdez11} was used in this analysis. 
				The key features of the datasets are presented in Table~\ref{table:multiclassdatasets}. 
				The IR is computed as the proportion of the number of the majority class examples to the number of minority class examples. 
				In this case, the class with maximum number of examples is the majority class, and the class with the minimum number of examples is the minority one. 
				We grouped the datasets according to their IRs using the group definitions suggested in Ref.~\citen{Alberto08}. 
				Datasets with low IR ($IR < 3$) are highlighted with dark gray, whereas datasets with medium IR ($3 < IR < 9$) are in light gray.
				
				\begin{table*}[!h]
					\centering
					\large
					\caption{Characteristics of the 26 multi-class imbalanced datasets taken from the Keel repository. Column \#E shows the number of instances in the dataset, column \#A the number of attributes (numeric/nominal), \#C shows the number of classes in the dataset, and column IR the imbalance ratio.}
					\label{table:multiclassdatasets}
					\resizebox{1\textwidth}{!}{
						\begin{tabular}{c c c c c c c c c c}
							\hline
							\textbf{Dataset} & \textbf{\#E} & \textbf{\#A} & \textbf{\#C} & \textbf{IR} & \textbf{Dataset} & \textbf{\#E} & \textbf{\#A} & \textbf{\#C} & \textbf{IR}\\
							\midrule
							\rowcolor[gray]{0.3}					
							\textcolor{white}{Vehicle} & \textcolor{white}{846} & \textcolor{white}{(18/0)} & \textcolor{white}{4} & \textcolor{white}{1.09} & \cellcolor{white} CTG & \cellcolor{white} 2126 & \cellcolor{white}  (21/0) & \cellcolor{white} 3 & \cellcolor{white}  9.40 \\
							\rowcolor[gray]{0.3}					
							\textcolor{white}{Wine} & \textcolor{white}{178} & \textcolor{white}{(13/0)} & \textcolor{white}{3} & \textcolor{white}{1.48} & \cellcolor{white} Zoo & \cellcolor{white}101 &\cellcolor{white} (16/0) &\cellcolor{white} 7 &\cellcolor{white} 10.25\\
							\rowcolor[gray]{0.3}
							\textcolor{white}{Led7digit} & \textcolor{white}{500} & \textcolor{white}{(7/0)} & \textcolor{white}{10} & \textcolor{white}{1.54} &\cellcolor{white} Cleveland & \cellcolor{white} 467 & \cellcolor{white} (13/0) & \cellcolor{white} 5 & \cellcolor{white} 12.62\\
							\rowcolor[gray]{0.3}					
							\textcolor{white}{Contraceptive} & \textcolor{white}{1473} & \textcolor{white}{(9/0)} & \textcolor{white}{3} & \textcolor{white}{1.89} & \cellcolor{white} Faults & \cellcolor{white} 1941 & \cellcolor{white} (27/0) & \cellcolor{white} 7 & \cellcolor{white} 14.05\\
							\rowcolor[gray]{0.3}
							\textcolor{white}{Hayes-Roth} & \textcolor{white}{160} & \textcolor{white}{(4/0)} & \textcolor{white}{3} & \textcolor{white}{2.10} & \cellcolor{white} Autos & \cellcolor{white} 159 & \cellcolor{white} (16/10) & \cellcolor{white} 6 & \cellcolor{white} 16.00\\
							\rowcolor[gray]{0.3}
							\textcolor{white}{Column3C} & \textcolor{white}{310} & \textcolor{white}{(6/0)} & \textcolor{white}{3} & \textcolor{white}{2.50} & \cellcolor{white} Thyroid & \cellcolor{white} 7200 & \cellcolor{white} (21/0) & \cellcolor{white} 3 & \cellcolor{white} 40.16\\
							\rowcolor[gray]{0.3}
							\textcolor{white}{Satimage} & \textcolor{white}{6435} & \textcolor{white}{(36/0)} & \textcolor{white}{7} &\textcolor{white}{2.45} & \cellcolor{white} Lymphography & \cellcolor{white} 148 & \cellcolor{white} (3/15) & \cellcolor{white} 4 & \cellcolor{white} 40.50\\
							\rowcolor[gray]{0.7}
							Laryngeal3 & 353 & (16/0) & 3 & 4.19 & \cellcolor{white} Abalone & \cellcolor{white} 4139 & \cellcolor{white} (7/1) & \cellcolor{white} 18 & \cellcolor{white} 45.93\\
							\rowcolor[gray]{0.7}
							New-thyroid & 215 & (5/0) & 3 & 5.00 & \cellcolor{white} Post-Operative & \cellcolor{white} 87 & \cellcolor{white} (1/7) & \cellcolor{white} 3 & \cellcolor{white} 62.00\\
							\rowcolor[gray]{0.7}
							Dermatology & 358 & (33/0) & 6 & 5.55  & \cellcolor{white} Wine-quality red & \cellcolor{white} 1599 & \cellcolor{white}(11/0) & \cellcolor{white} 11 &  \cellcolor{white}68.10\\
							\rowcolor[gray]{0.7}
							Balance & 625 & (4/0) & 3 & 5.88 & \cellcolor{white} Ecoli & \cellcolor{white} 336 & \cellcolor{white} (7/0) &\cellcolor{white} 8 &\cellcolor{white} 71.50\\
							\rowcolor[gray]{0.7}
							Flare & 1066 & (0/11) & 6 & 7.70 & \cellcolor{white} Page-blocks &\cellcolor{white} 5472 & \cellcolor{white} (10/0) & \cellcolor{white} 5 & \cellcolor{white} 175.46\\
							\rowcolor[gray]{0.7}
							Glass & 214 & (9/0) & 6 & 8.44 & \cellcolor{white} Shuttle & \cellcolor{white} 2175 & \cellcolor{white} (9/0) & \cellcolor{white} 5 &  \cellcolor{white}853.00\\
							
							\hline
						\end{tabular}}
					\end{table*} 

\subsection{Experimental setup}

Experiments were performed using the DESLib\cite{deslib}, and the results were obtained with a $5\times 2$ stratified cross-validation. 
Performance evaluation is conducted using multi-class generalizations of the AUC, F-measure and G-mean, as the standard classification accuracy is not suitable for imbalanced learning~\cite{Pastor15}. 
In particular, the multi-class generalization of the AUC used in this work was the one from Ref.~\citen{hand2001simple}, which performs an estimation of the AUC for each pair of classes and then returns their averaged score. 
The generalization of the F-measure followed the weighted mean approach, which calculates the metric for each class individually and then combines them using a weighted sum, with the proportion of samples from each corresponding class as the weights. 
Lastly, the G-mean score was obtained in this work as a higher root of the product of sensitivity for each class.

The pool size for all ensemble techniques was set to 100. 
The classifier used as a base classifier in all ensembles was the decision tree implementation from the Python library Scikit-learn\cite{sklearn}, which uses the CART\cite{cart} algorithm. 
Here, the decision trees were used without pruning and collapsing as recommended in Ref.~\citen{Pastor15}. 
However, the minimum impurity decrease was set to 0.05 to function as an early stop in the training phase in order to improve the probability estimation of the unpruned trees.

\begin{table}[!h]
	\centering
	\caption{Preprocessing methods used for classifier pool generation. 
	\textit{Note.} Reprinted from ``A study on combining dynamic selection and data preprocessing for imbalance learning," by A. Roy, R. Cruz, R. Sabourin and G. Cavalcanti, 2018,\textit{ Neurocomputing}, \textit{286}, 179--192. Copyright 2018 by Elsevier.
	}\label{tbl:EnsembleMethods}
	\resizebox{1\textwidth}{!}{  
	\begin{tabular}{lll}
		\hline
		\multicolumn{3}{l}{Bagging based methods} \\
		\hline
		Abbr. & Name & Description \\
		\hline
		Ba & Bagging & Bagging without preprocessing\\
		Ba-RM100 & Bagging+RAMO 100\%& RAMO to double the minority class \\
		Ba-RM & Bagging+RAMO & RAMO to make equal size for both classes \\
		Ba-SM100 & Bagging+SMOTE 100\% & SMOTE to double the minority class \\
		Ba-SM & Bagging+SMOTE & SMOTE to make equal size for both classes \\
		Ba-RB & Bagging+RB & RB to randomly balance the two classes \\
		\hline
	\end{tabular}}
\end{table}

All preprocessing techniques were combined with Bagging during the pool generation phase. 
Table \ref{tbl:EnsembleMethods} lists such combinations. 
The preprocessing techniques, RAMO and SMOTE, have user-specified parameters. 
In the case of RAMO, we used $k_1 = 10$, $k_2 = 5$ and $\alpha = 0.3$. 
For SMOTE and RB, the number of nearest neighbors was 5. 
These parameter settings were adopted from Ref.~\citen{Pastor15}. 
Finally, for all the dynamic selection methods, we used 7 nearest neighbors to define the region of competence as in Ref.~\citen{Cruz15} and Ref.~\citen{CRUZ2018195}.

\begin{figure}[!h]

			\centering
			\includegraphics[width=0.75\textwidth]{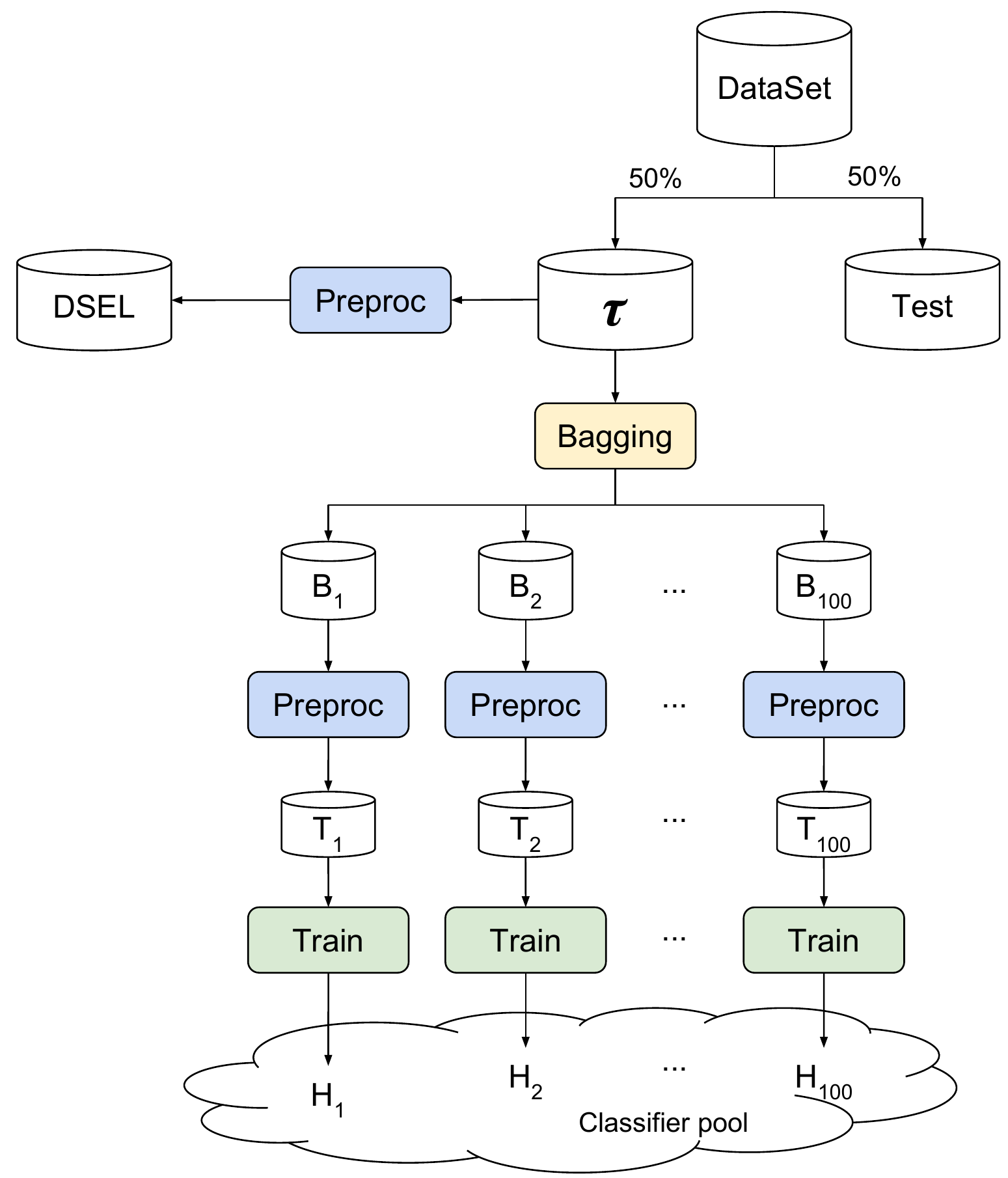}\\
			\caption{The framework for training base classifiers and to prepare a DSEL for testing. Here, $\vec{\tau}$ is the training data derived from the original dataset, $B_i$ is the dataset generated from the $i$th Bagging iteration, $T_i$ is the dataset produced by preprocessing (\textit{Preproc}) $B_i$ and $H_i$ is the $i$th base classifier.
		Reprinted from ``A study on combining dynamic selection and data preprocessing for imbalance learning," by A. Roy, R. Cruz, R. Sabourin and G. Cavalcanti, 2018, \textit{Neurocomputing}, \textit{286}, 179--192. Copyright 2018 by Elsevier.}\label{fig:OneFold}
		\end{figure}

The complete framework for a single replication is presented in Figure \ref{fig:OneFold}. 
The original dataset was divided into two equal halves. One of them was set aside for testing, while the other half was used to train the base classifiers and to derive the dynamic selection set. 
Let us now highlight the process of setting up the DSEL. 
Here, instead of dividing the training set, we augment it using the data preprocessing, to create DSEL. 
Moreover, the Bagging method is applied to the training set, generating a bootstrap with 50\% of the data. 
Then, the preprocessing method is applied to each bootstrap, and the resulting dataset is used to generate the pool of classifiers. 
Since we considered a single training dataset, the DSEL dataset has an overlap with the datasets used during Bagging iterations. 
However, the randomized nature of the preprocessing methods allows the DSEL not to be exactly the same as the training datasets, thus avoiding overfitting issues. 
Moreover, the use of overlapping training and validation sets was shown to improve the performance of classifier ensembles in comparison with disjoints datasets in Ref.~\citen{dietrich2003decision}.

\subsection{Results according to data preprocessing method}

In this section, we compare the performance of each preprocessing method with respect to each ensemble technique. 
Tables~\ref{table:rank-preproc}a, \ref{table:rank-preproc}b and \ref{table:rank-preproc}c show the average rank for the AUC, F-measure and G-mean, respectively. 
The best average rank is in bold. We can see that the RM and RM100 obtained the best results. 
Furthermore, the configuration using only Bagging always presented the greatest average rank for the AUC and G-mean. 

\begin{table}[!htb]
\centering 
\caption{Average rankings according to (a) AUC, (b) F-measure and (c) G-mean. Methods in brackets are statistically equivalent to the best one.}
\label{table:rank-preproc}
\subfloat[]{   
	\resizebox{0.95\textwidth}{!}{  
	\begin{tabular}{c c c c c c c}  
	\hline  
 
	Algorithm & Bagging & RM & RM100 & SM & SM100 & RB \\ \hline
  
	STATIC			& 4.96 & {\bf 2.39} & [3.39] & [3.25] & 4.21 & [2.79] \\
	RANK			& 5.07 & {\bf 2.54} & [2.89] & [3.04] & [3.57] & 3.89 \\
	LCA				& 4.93 & {\bf 2.14} & 3.57 & [2.79] & 3.39 & 4.18 \\
	F-LCA			& 4.68 & {\bf 2.43} & [3.57] & [2.96] & [3.29] & 4.07 \\
	MCB				& 4.98 & {\bf 2.66} & [2.91] & [2.88] & [3.70] & 3.88 \\
	F-MCB			& 5.04 & {\bf 2.36} & [3.30] & [3.07] & [3.38] & 3.86 \\
	KNE				& 5.61 & [2.68] & {\bf 2.66} & [3.45] & [3.79] & [2.82] \\
	F-KNE			& 5.64 & [2.71] & {\bf 2.64} & [3.50] & [3.79] & [2.71] \\
	KNU				& 5.70 & [2.48] & [3.14] & [3.23] & 4.23 & {\bf 2.21} \\
	F-KNU			& 5.71 & [2.41] & [3.20] & [3.27] & 4.12 & {\bf 2.29} \\
	DES-KNN			& 5.96 & {\bf 2.09} & [3.14] & 3.39 & 4.05 & [2.36] \\
	F-DES-KNN		& 5.96 & {\bf 2.20} & [3.07] & 3.41 & 3.96 & [2.39] \\
	DESP			& 5.88 & [2.45] & [2.89] & [3.38] & 3.98 & {\bf 2.43} \\
	DES-RRC			& 5.89 & {\bf 2.43} & [2.96] & [3.36] & 3.86 & [2.50] \\
	META-DES		& 5.89 & [2.59] & [2.98] & [3.20] & 3.77 & {\bf 2.57} \\
	\hline  
 
	\end{tabular}}}
\end{table} 
	
	\begin{table}[!htbp]
\ContinuedFloat
\captionsetup{position=top}
\subfloat[]{   
\resizebox{0.95\textwidth}{!}{  
\begin{tabular}{c c c c c c c}  
\hline  
 
Algorithm & Bagging & RM & RM100 & SM & SM100 & RB \\ \hline  

STATIC			& [3.54] & [4.27] & {\bf 2.86} & [3.21] & [3.25] & [3.88] \\
RANK			& [3.43] & 4.20 & {\bf 2.55} & [3.16] & [2.96] & 4.70 \\
LCA				& [3.32] & 3.89 & [3.20] & [2.93] & {\bf 2.41} & 5.25 \\
F-LCA			& [3.04] & 4.00 & [3.18] & [3.14] & {\bf 2.46} & 5.18 \\
MCB				& [3.36] & 4.32 & {\bf 2.68} & [3.04] & [3.11] & 4.50 \\
F-MCB			& [3.32] & 4.05 & {\bf 2.46} & [3.30] & [3.20] & 4.66 \\
KNE				& [3.70] & 4.00 & {\bf 2.54} & [3.54] & [3.02] & 4.21 \\
F-KNE			& [3.77] & 4.05 & {\bf 2.50} & [3.50] & [3.00] & 4.18 \\
KNU				& [3.75] & [4.14] & {\bf 2.93} & [3.21] & [3.32] & [3.64] \\
F-KNU			& [3.61] & [4.20] & {\bf 2.98} & [3.21] & [3.43] & [3.57] \\
DES-KNN			& 4.11 & 3.89 & {\bf 2.36} & [3.11] & [3.21] & 4.32 \\
F-DES-KNN		& 4.07 & 3.86 & {\bf 2.50} & [3.14] & [3.36] & 4.07 \\
DESP			& [3.57] & [4.18] & {\bf 2.79} & [3.23] & [3.27] & [3.96] \\
DES-RRC			& [3.57] & [4.25] & {\bf 2.86} & [3.32] & [3.18] & [3.82] \\
META-DES		& [4.00] & [3.96] & [2.91] & {\bf 2.86} & [3.34] & [3.93] \\
\hline  
 	\end{tabular}} }
\\
\subfloat[]{   
\resizebox{0.95\textwidth}{!}{  
\begin{tabular}{c c c c c c c}  
\hline  
 
Algorithm & Bagging & RM & RM100 & SM & SM100 & RB \\ \hline  
 
STATIC			& 5.54 & {\bf 2.46} & [3.29] & [3.43] & 3.75 & [2.54] \\
RANK			& 5.59 & {\bf 2.30} & [3.14] & [3.14] & 3.64 & [3.18] \\
LCA				& 5.32 & {\bf 2.04} & 3.57 & 3.29 & 3.29 & 3.50 \\
F-LCA			& 5.14 & {\bf 2.18} & 3.64 & [3.21] & [3.20] & 3.62 \\
MCB				& 5.48 & {\bf 2.32} & [3.12] & [3.00] & 4.07 & [3.00] \\
F-MCB			& 5.71 & {\bf 2.14} & [2.79] & [3.07] & 4.21 & [3.07] \\
KNE				& 5.75 & {\bf 2.29} & [2.79] & 3.64 & 3.82 & [2.71] \\
F-KNE			& 5.75 & {\bf 2.30} & [2.79] & 3.59 & 3.86 & [2.71] \\
KNU				& 5.61 & [2.46] & 3.39 & 3.57 & 3.86 & {\bf 2.11} \\
F-KNU			& 5.75 & [2.43] & 3.45 & 3.34 & 3.93 & {\bf 2.11} \\
DES-KNN			& 5.75 & {\bf 2.18} & [3.00] & 3.32 & 4.04 & [2.71] \\
F-DES-KNN		& 5.82 & {\bf 2.32} & [3.02] & [3.32] & 3.84 & [2.68] \\
DESP			& 5.61 & [2.64] & [3.30] & [3.25] & 3.89 & {\bf 2.30} \\
DES-RRC			& 5.54 & {\bf 2.43} & [3.39] & [3.32] & 3.79 & 2.54] \\
META-DES		& 5.82 & [2.57] & [3.29] & 3.46 & 3.64 & {\bf 2.21} \\
\hline  
\end{tabular}} }
\end{table} 

The Finner's\cite{Finner93} step-down procedure was conducted at a $95\%$ significance level to identify all methods that were equivalent to the best ranked one. 
The analysis demonstrates that considering the AUC and G-mean, the result obtained using preprocessing techniques is always statistically better when compared to using only Bagging. 
The same was not observed for the F-measure, probably because we used the weighted multi-class generalization of this measure, in which the classes with more samples have greater weights, and thus influence, in the calculation of the final F-measure score. 

It can also be observed from Table \ref{table:rank-preproc} that, for all metrics, the best preprocessing method for any given DS technique was also the best one for the same technique within the FIRE-DES framework. 
This shows that there is not a single preprocessing technique that particularly favors the FIRE-DES scheme. Rather, the impact of the preprocessing procedure over the DS techniques is hardly changed by using them within the framework.

\begin{figure*}[!hp]
	\centerline{
	\subfloat[AUC]{\includegraphics[width=1\textwidth]{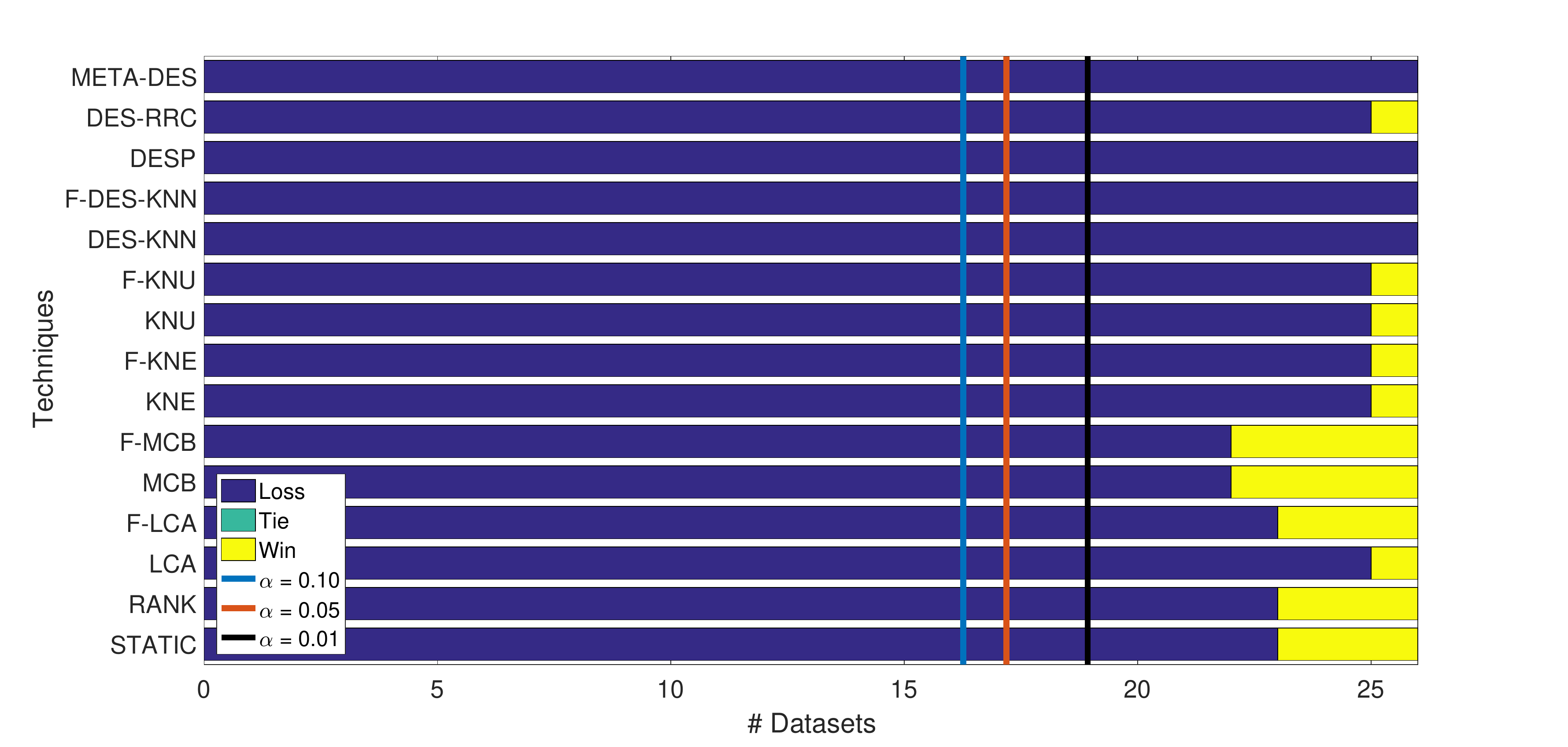}}  } 
	\centerline{
	\subfloat[F-measure]{\includegraphics[width=1\textwidth]{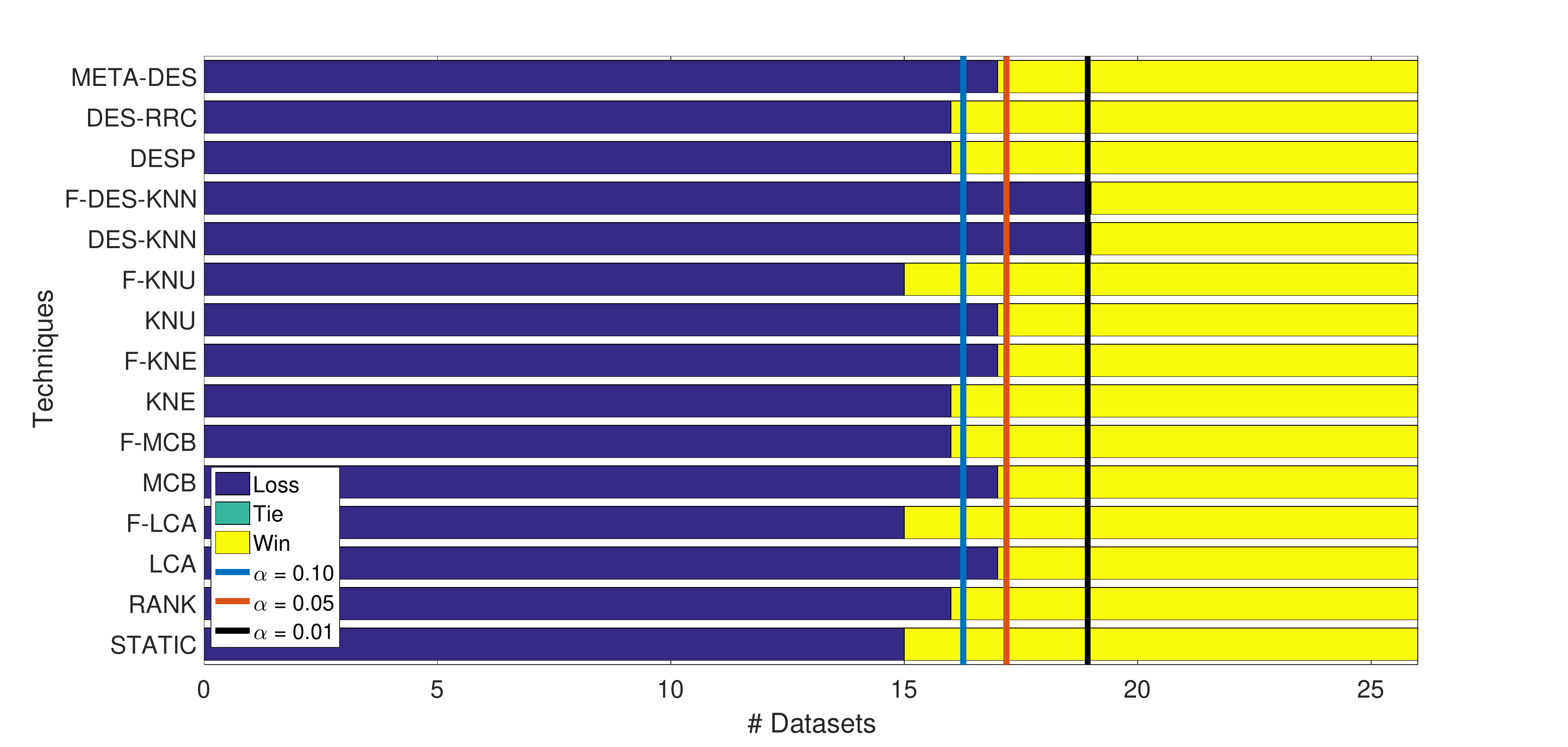}}}
	\centerline{	
	\subfloat[G-mean]{\includegraphics[width=1\textwidth]{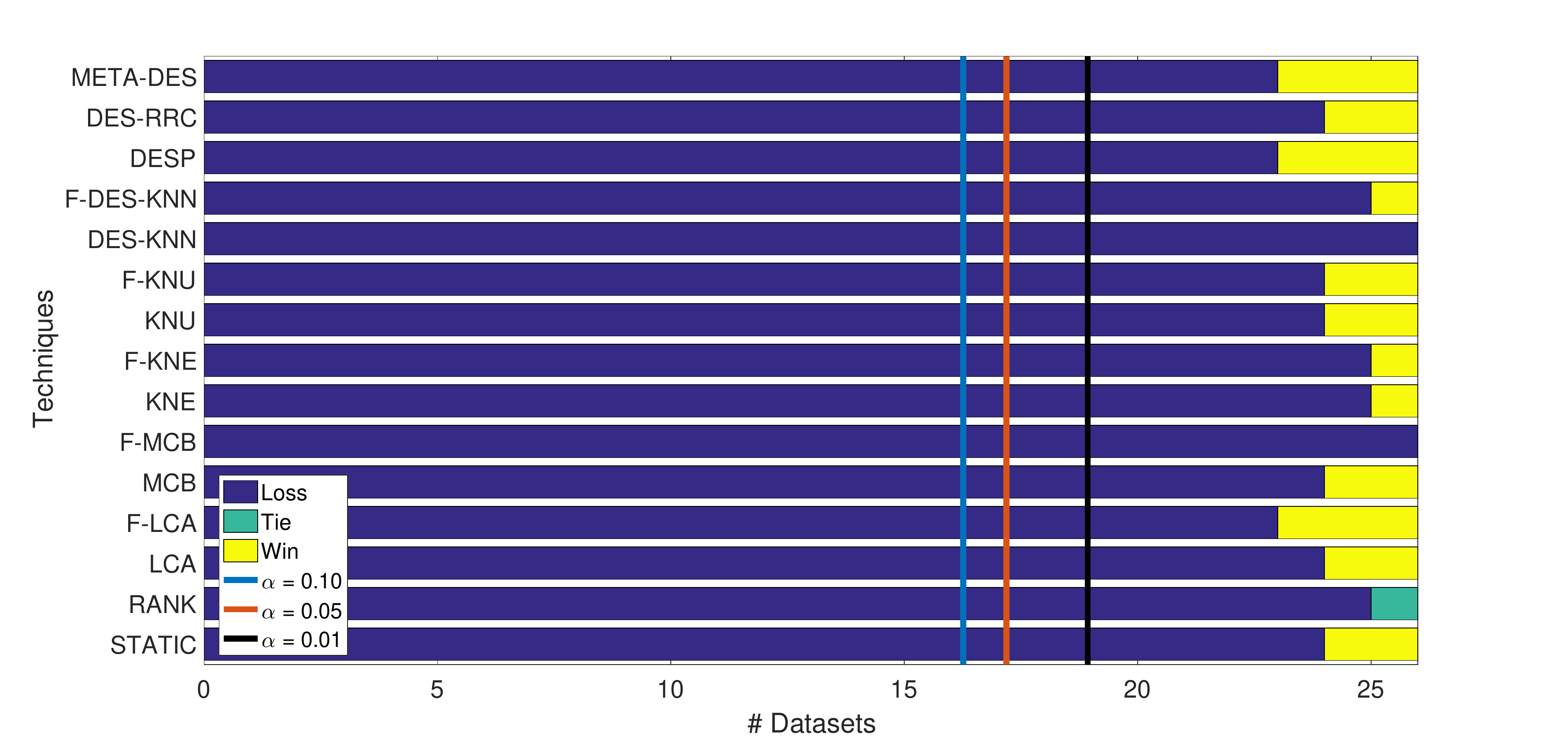}}   }
	
	\caption{Sign test computed over the wins, ties and losses. The vertical lines represents the critical value for at a significance level $\alpha = \left\lbrace 0.1, 0.05, 0.01 \right\rbrace $.}
	\label{fig:wintielosDES}
\end{figure*}

Moreover, we conducted a pairwise comparison between each ensemble methods using data preprocessing with the same methods using only Bagging (baseline). 
For the sake of simplicity, only the best data preprocessing for each technique was considered (i.e., the best result of each row of Table \ref{table:rank-preproc}). 
The pairwise analysis is conducted using the Sign test, calculated on the number of wins, ties, and losses obtained by each method using preprocessing techniques, compared to the baseline. 
The results of the Sign test is presented in Figure~\ref{fig:wintielosDES}.

The Sign test demonstrated that the data preprocessing significantly improved the results of these techniques according to the AUC and G-mean. 
Considering these two metrics, all techniques obtained a significant number of wins for a significance level $\alpha = 0.01$. 
For the F-measure, nearly half of the techniques presented a significant number of wins for $\alpha = 0.10$. 
Hence, the results obtained demonstrate that data preprocessing techniques indeed play an important role when dealing with multi-class imbalanced problems. 

Furthermore, DS techniques are more benefited from the application of data preprocessing (i.e., presented a higher number of wins). 
This result can be explained by the fact the data preprocessing techniques are applied in two stages. 
First, it is used in the ensemble generation stage in order to generate a diverse pool of classifiers. 
Then, they are also used in order to balance the distribution of the dynamic selection dataset for the estimation of the classifiers' competences.

Figure \ref{fig:wintielosDES} also shows that, for the DS techniques within the FIRE-DES framework, the number of wins is almost always less or equal to the number of wins from the same DS technique by itself, for all metrics. 
This suggests that, as opposed to the two-class problems' case, the FIRE-DES scheme does not yield an improvement in performance for the DS techniques for multi-class imbalanced problems.
\newpage

\subsection{Dynamic selection vs static combination}

In this experiment we compare the performance of the dynamic selection approaches versus static ones. 
For each technique, the best performing data preprocessing technique is selected (i.e., best result from each row of Table \ref{table:rank-preproc}). 
Then, new average ranks are calculated for these methods. Table \ref{table:ranks} presents the average rank of the top techniques according to each metric.

Based on the average ranks, we can see that all DES techniques presented a lower average rank when compared to that of the static combination for the three performance measures. 
In fact, the performance of most DES techniques was significantly better than the static combination's for both the AUC and the G-mean. 
Hence, DES techniques are suitable for dealing with multi-class imbalance. 
The DCS techniques, however, obtained a greater average rank in comparison with the static combination for the F-measure and G-mean, suggesting that they may not be fit for handling multi-class imbalanced problems. 
Furthermore, the DS techniques within the FIRE-DES framework yielded, in almost every case, a greater average rank than the same technique by itself, further suggesting that the FIRE-DES may be unsuitable for multi-class imbalanced classification.
					
						\begin{table}[!htb]
							\renewcommand\thetable{4}
							\centering
							\caption{Average ranks for the best ensemble methods. (a) According to AUC, (b) according to F-measure and (c) according to G-mean. Results that are statistically equivalent to the best one are in brackets.}
							\label{table:ranks}
							\resizebox{1\textwidth}{!}{
								\begin{tabular}{@{\extracolsep{9pt}}@{\kern\tabcolsep}l@{}r@{}l@{}r@{}l@{}r@{\kern\tabcolsep}}
									\toprule
									\multicolumn{2}{l}{(a) AUC} & \multicolumn{2}{l}{(b) F-measure} & \multicolumn{2}{l}{(c) G-mean} \\
									\cmidrule{1-2}\cmidrule{3-4}\cmidrule{5-6}
									Methods & Rank & Method & Rank & Method & Rank \\
									
									\midrule
									Ba-RB+DESP		& {\bf 2.64} & Ba-RM100+KNU			&   {\bf 4.73}  & Ba-RB+KNU			&  {\bf 4.95} \\
									Ba-RB+DES-RRC	&   [3.16] 	 & Ba-RM100+DES-RRC		&   [5.30] 		& Ba-RB+META-DES   	&  [5.20] \\
									Ba-RB+META-DES  &   [3.25]	 & Ba-RM100+DES-KNN		&   [5.54] 		& Ba-RB+F-KNU		&  [5.41] \\
									Ba-RM+DES-KNN	&   [4.73] 	 & Ba-SM+META-DES   	&   [6.05] 		& Ba-RB+DESP		&  [6.34] \\
									Ba-RB+KNU		&   [4.80] 	 & Ba-RM100+DESP		&   [6.18] 		& Ba-RM+F-DES-KNN 	&  [6.52] \\
									Ba-RM+F-DES-KNN	&   [5.02] 	 & Ba-RM100+F-DES-KNN 	&   [6.29] 		& Ba-RM+DES-KNN		&  [6.54] \\
									Ba-RB+F-KNU		&   6.25 	 & Ba-RM100+KNE			&   [6.43] 		& Ba-RM+KNE			&  [6.79] \\
									Ba-RM100+KNE	&   8.30 	 & Ba-RM100+F-KNE		&   [7.05] 		& Ba-RM+DES-RRC		&  [7.12] \\
									Ba-RM100+F-KNE	&   8.73 	 & Ba-RM100+F-KNU		&   [7.41] 		& Ba-RM+F-KNE		&  [7.21] \\
									Ba-RM+F-MCB		&   11.80 	 & Ba-RM100    			&  [7.41] 		& Ba-RM    			& 7.79 \\
									Ba-RM+MCB		&   11.84 	 & Ba-RM100+MCB			&   10.54 		& Ba-RM+MCB			&  9.57 \\
									Ba-RM+LCA    	&   11.96 	 & Ba-RM100+F-MCB		&   10.86 		& Ba-RM+F-MCB		&  10.27 \\
									Ba-RM    		&   12.14 	 & Ba-RM100+RANK   		&   11.36 		& Ba-RM+LCA    		&  11.52 \\
									Ba-RM+F-LCA		&   12.50 	 & Ba-SM100+LCA    		&   12.02 		& Ba-RM+RANK   		&  12.14 \\
									Ba-RM+RANK   	&   12.86 	 & Ba-SM100+F-LCA		&   12.84 		& Ba-RM+F-LCA		&  12.64 \\
									%
									
									\bottomrule
								\end{tabular}}
							\end{table}

\begin{figure*}[!htbp]
	\centerline{
	\subfloat[AUC]{\includegraphics[width=1\textwidth]{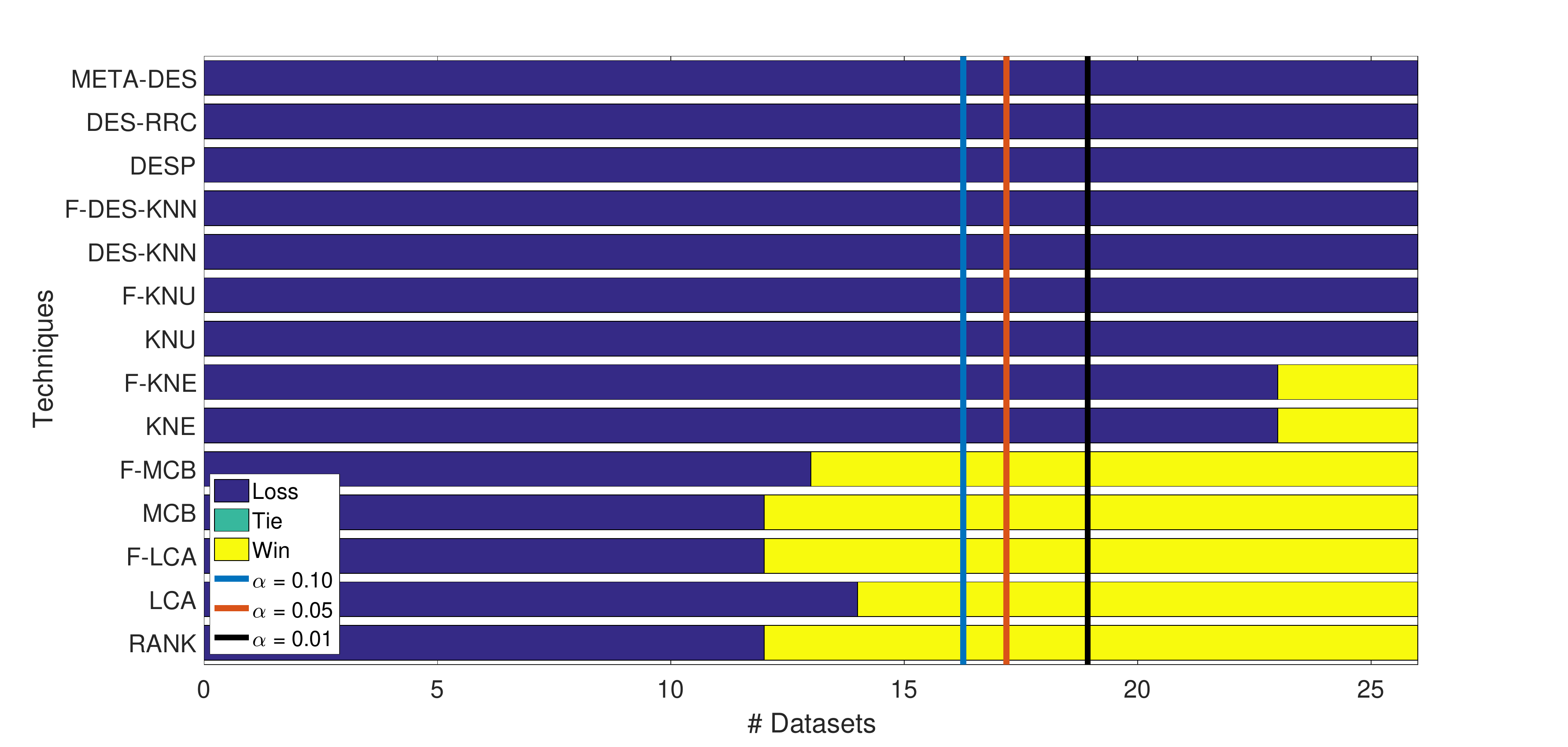}}  } 
	\centerline{
	\subfloat[F-measure]{\includegraphics[width=1\textwidth]{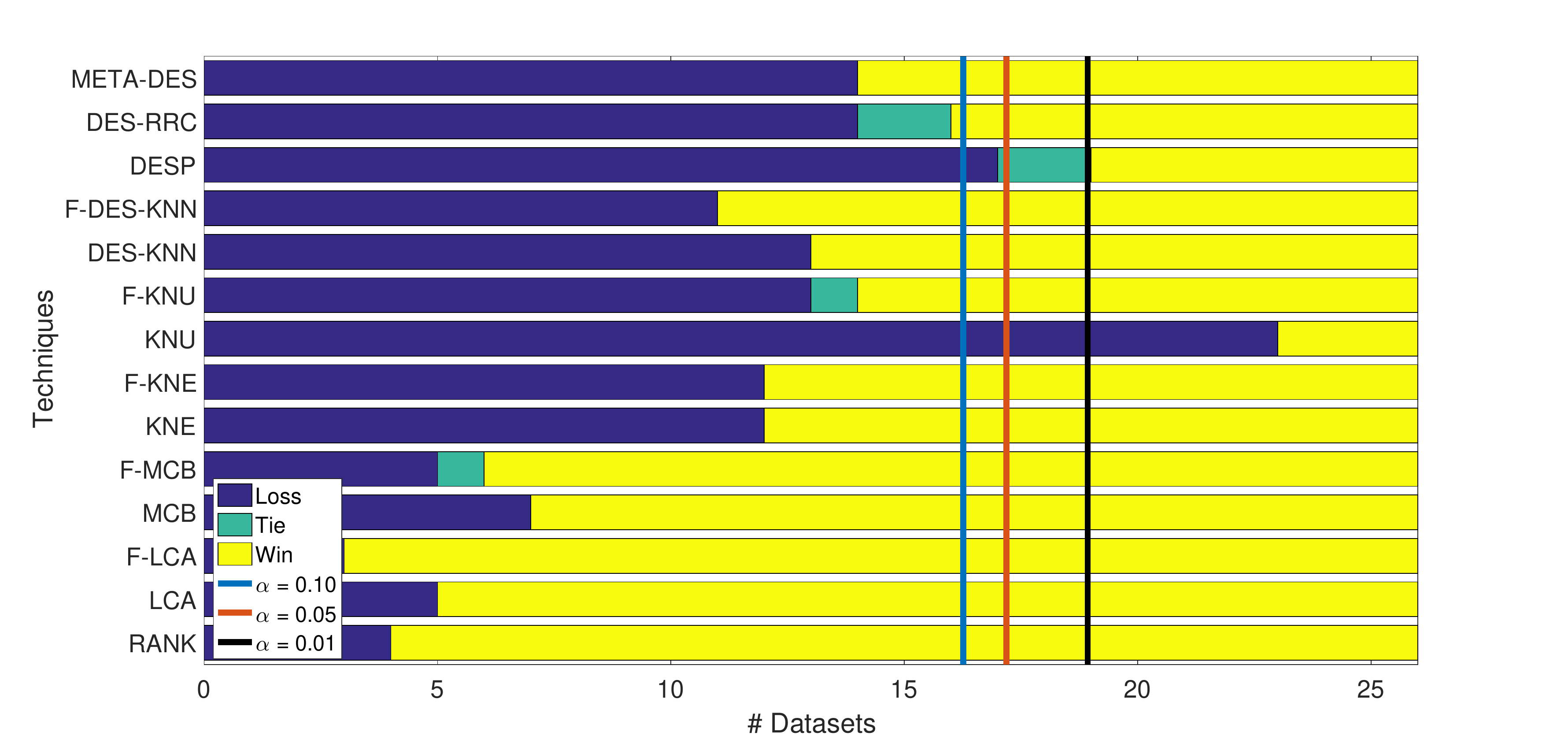}}}
	\centerline{	
	\subfloat[G-mean]{\includegraphics[width=1\textwidth]{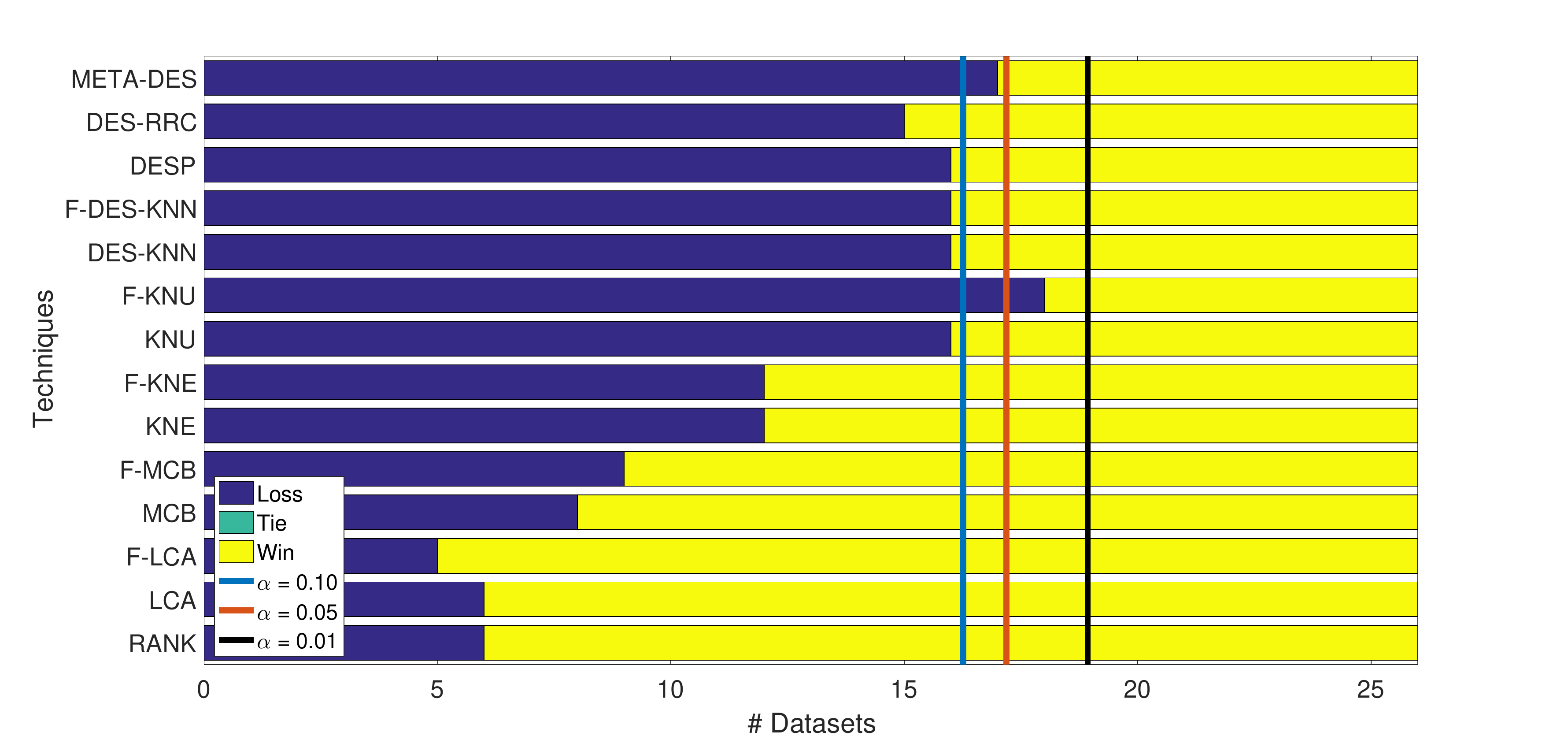}}   }
	
	\caption{Sign test computed over the wins, ties and losses. The vertical lines represents the critical value for at a significance level $\alpha = \left\lbrace 0.1, 0.05, 0.01 \right\rbrace $.}
	\label{fig:wintielosDES-Static}
	
\end{figure*}

We also performed a pairwise comparison between the best data preprocessing for each DS technique and the best data preprocessing for the static combination, that is, between each row from Tables \ref{table:ranks}a, \ref{table:ranks}b and \ref{table:ranks}c and the row in the same table that corresponds to Bagging. 
Figure \ref{fig:wintielosDES-Static} shows the results of the Sign test for each metric.  

It can be observed that, for the AUC, all DES techniques performed significantly better than Bagging for $\alpha = 0.01$. 
The DCS techniques, on the other hand, obtained fewer wins in comparison with the static combination, save for LCA. 
For the F-measure, only the KNU and the DESP yielded a significantly superior performance compared to Bagging. 
However, nearly half of the DES techniques obtained more wins than losses in the Sign test for this metric. 
As in the AUC case, the DCS techniques yielded a poorer performance in comparison with the static combination.
For the G-mean, the F-KNU and the DESP obtained a significantly superior performance in comparison with Bagging for $\alpha = 0.01$ and $\alpha = 0.10$, respectively. 
Moreover, most DES techniques also yielded a greater number of wins for this metric, while all DCS techniques obtained much fewer wins in comparison with the static combination.

Overall, the DES techniques performed better than the static combination, further showing its suitability for handling multi-class imbalanced problems.  
The DCS techniques, however, obtained a poorer performance in comparison with Bagging for the three metrics, which suggests it may be unsuitable for multi-class imbalanced classification.
As for the FIRE-DES, save for a few exceptions, the DS techniques within the framework obtained the same or fewer number of wins in comparison with the same technique on its own, which again suggests that it may not be fit for dealing with multi-class imbalance.

\section{Conclusion}

In this work, we conducted a study on dynamic ensemble selection and data preprocessing for solving the multi-class imbalanced problems. 
A total of fourteen dynamic selection schemes and five preprocessing techniques were evaluated in this experimental study.

Results obtained over 26 multi-class imbalanced problems demonstrate that the dynamic ensemble selection techniques studied obtained a better result than static ensembles based on AUC, F-measure and G-mean. 
Moreover, the use of data preprocessing significantly improves the performance of DS and static ensembles. 
In particular, the RAMO technique presented the best overall results. 
Furthermore, DS techniques seems to benefit more of data preprocessing methods since they are applied not only to generate the pool of classifiers but also to edit the distribution of the dynamic selection dataset.

Future works would involve the definition of new pre-processing techniques specific to deal with multi-class imbalance as well as the definition of cost-sensitive dynamic selection techniques to handle multi-class imbalanced problems.

\section*{Acknowledgments}

The authors would like to thank the Brazilian agencies CAPES (Coordena\c{c}\~{a}o de Aperfei\c{c}oamento de Pessoal de N\'{i}vel Superior), CNPq (Conselho Nacional de Desenvolvimento Cient\'{i}fico e Tecnol\'{o}gico) and FACEPE (Funda\c{c}\~{a}o de Amparo \`{a} Ci\^{e}ncia e Tecnologia de Pernambuco) and the Canadian agency NSERC (Natural Sciences and Engineering Research Council of Canada).


\bibliographystyle{ws-ijprai}
\bibliography{ThesisBib}


%
%
%

\end{document}